\documentclass[11pt]{article}

\usepackage[preprint]{acl}

\usepackage{times}
\usepackage{latexsym}

\usepackage[T1]{fontenc}

\usepackage[utf8]{inputenc}

\usepackage{microtype}

\usepackage{inconsolata}

\usepackage{graphicx}

\usepackage{layouts}

\usepackage{booktabs}
\usepackage{tabularray}

\usepackage{amsmath, amssymb}

\usepackage{etoc}

\usepackage[x11names]{xcolor}

\usepackage{pdflscape}

\usepackage{lipsum}

\usepackage{subcaption}

\usepackage{xspace}

\usepackage{diagbox}

\usepackage{siunitx}

\usepackage{enumitem}

\usepackage{simpleicons}

\usepackage{worldflags}

\newcommand{\tm}{\fcolorbox{black}{LightPink1}{\texttt{TM}}\xspace}
\newcommand{\tg}{\fcolorbox{black}{LightBlue1}{\texttt{TG}}\xspace}
\newcommand{\cm}{\fcolorbox{black}{PaleGreen1}{\texttt{CM}}\xspace}

\newcommand{\parheading}[1]{
    \vspace{0.5em}
    \par
    \noindent
    \textbf{#1}
}

\newcommand{\prism}{\textsc{prism}\xspace}
\newcommand{\community}{\textsc{community}\xspace}

\title{What do Reward Models Memorize?}

\author{ \bf Ivo Verhoeven\textsuperscript{\includegraphics[height=1.05em]{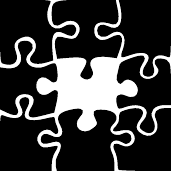}},
Pushkar Mishra\textsuperscript{\includegraphics[height=1.05em]{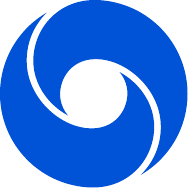}},
Ekaterina Shutova\textsuperscript{\includegraphics[height=1.05em]{figures/illc_logo.pdf}}\\
\includegraphics[height=0.7em]{figures/illc_logo.pdf}
ILLC, University of Amsterdam\\
\includegraphics[height=0.7em]{figures/deepmind_logo.pdf}
Google DeepMind, London \\
{\small \href{mailto:i.o.verhoeven@uva.nl}{\texttt{i.o.verhoeven@uva.nl}}, \texttt{pm576@cantab.ac.uk}, \texttt{e.shutova@uva.nl}}
}

\hypersetup{
    pdfauthor={Ivo Verhoeven, Pushkar Mishra, Ekaterina Shutova},
    pdftitle={What do Reward Models Memorize?},
}

\begin{document}
\maketitle
\etocdepthtag.toc{mtchapter} \etocsettagdepth{mtchapter}{subsection}
\etocsettagdepth{mtappendix}{none}




\begin{abstract}
    
    This paper studies what discriminatively trained reward models (RMs) memorize by measuring counterfactual memorization on two human preference datasets. We show that RMs 1) misallocate memorization to easy, high margin preference pairs, 2) memorize dataset-specific shortcuts (e.g., model identity, user sampling strategy), and 3) overgeneralize simple heuristic correlates of human preference (e.g., length, compliance) when confronted with unseen preference pairs. Overall, our findings indicate that discriminative training of RMs from human preference data results in biased RMs not yet capable of judging response quality in context-dependent scenarios.
    
    \simpleicon{git} \href{https://github.com/ioverho/rm-shortcuts}{https://github.com/ioverho/rm-shortcuts}
\end{abstract}

\section{Introduction}

Converting general purpose Large Language Models (LLMs) into human-preference aligned chat models requires maximizing a safety constrained user utility function. The predominant optimization paradigm for this is Reinforcement Learning from Human Feedback (RLHF) \cite{askellGeneralLanguageAssistant2021, ouyangTrainingLanguageModels2022, openaiIntroducingChatGPT2024}. Learning algorithms belonging to this paradigm typically use a Reward Model (RM) as a proxy for the intangible, inconsistent, and often inaccessible true human utility function. An RM takes as input a user prompt and an LLM response, and outputs a scalar quality estimate. This RM can be explicit (e.g., PPO \cite{schulmanProximalPolicyOptimization2017}) or implicit (e.g., DPO \cite{rafailovDirectPreferenceOptimization2023}), but its guidance is crucial for the development of maximally helpful yet minimally harmful chat models.

The use of an RM as a proxy for the true human preference function comes with serious risks if the RM is misspecified. For example, the RM might overfit to the training data and memorize dataset-dependent heuristic shortcut features that correlate but do not cause human preference. This can induce \textit{reward hacking}\footnote{Also known as reward overoptimization, specification gaming, Goodharting} during RLHF \cite{skalseDefiningCharacterizingReward2022, gaoScalingLawsReward2023}. This leads to LLM output maximizing RM scores without increasing user utility, having never learned the underlying causes of human preference.

Despite RLHF-optimized chat models being deemed safe enough for widespread public release, current research frequently reports LLM output behaviors indicative of reward hacking an overfit RM. For example, model responses are often overly verbose \cite{saitoVerbosityBiasPreference2023, singhalLongWayGo2024}; too affirming of a user's subjective experience and framing (i.e., sycophancy) \cite{wangWhenTruthOverridden2026, chengELEPHANTMeasuringUnderstanding2025}; too dependent on vague generalities \cite{sharmaUnderstandingSycophancyLanguage2024, wangWhenTruthOverridden2026}; reliant on discriminatory stereotypes or social prejudices \cite{wangRewardHackingCausal2025}; excessively stylized or formatted \cite{bharadwajFlatteryFluffFog2025}; etc.

The problem of rectifying these behaviors in LLM-based chat models has received a great deal of attention, yet the source of these behaviors remains unclear. If it is due to a misspecified, dataset-specific RM, what exactly is it that RMs memorize from a dataset? 

Memorization in and of itself does not negatively impact model performance; in some cases, memorization is beneficial to generalization \cite{anagnostidisCuriousCaseBenign2022, carliniQuantifyingMemorizationNeural2022, bayatPitfallsMemorizationWhen2024}, whereas in others it is a necessary condition for learning \cite{feldmanDoesLearningRequire2020, attiasInformationComplexityStochastic2024}. Regardless, some degree of memorization should be expected in LLMs \cite{mahdaviMemorizationCapacityMultihead2024, morrisHowMuchCan2026}. We are specifically interested in finding instances of RM memorization that negatively impact performance on difficult, context-dependent preference comparisons, or reward the above listed misaligned behaviors.

To do so, this paper studies RM memorization through the lens of counterfactual memorization (\cm)
\cite{feldmanDoesLearningRequire2020, zhengEmpiricalStudyMemorization2022,
zhangCounterfactualMemorizationNeural2023, dankersMemorisationCartographyMapping2023}. We associate \cm with a set of hypothesis- and data-driven features found in two human preference datasets. From our findings, we propose \num{3} memorization patterns induced by discriminative training of RMs. Specifically, we find that RMs 1) misallocate memorization to easy, large human preference margin response pairs; 2) can and do memorize dataset-specific features not causally related to human preference (e.g., response model idiosyncracies or rater metadata); 3) overgeneralize difference in simple heuristic features when presented with unseen preference pairs.

Altogether, we find that discriminative RM training yields models that memorize the wrong data and are overly dependent on dataset-specific shortcuts. 


\section{Related Work}

For the safe deployment of LLM based chat models, their behavior must be aligned to the preferences of judges capable of weighing conflicting principles against each other consistently \cite{buylAIAlignmentYour2025}. Since these judges are often inaccessible, many RLHF pipelines use some form of RM as a proxy. A litany of prior work, however, has shown that RMs do not do this balancing of principles.

In some cases, this is due to RMs overinflating the importance of surface-level correlates of human preference, i.e., overgeneralization. Examples include the dependence on differences in length \cite{saitoVerbosityBiasPreference2023, singhalLongWayGo2024, huangPosthocRewardCalibration2024}, formatting \cite{zhangListsEmojisHow2025}, style \cite{bharadwajFlatteryFluffFog2025}, and recently model identity \cite{zhuCHARMCalibratingReward2026}.

Prior work has also shown RMs to be insensitive to features \textit{that do} cause human preference \cite{shenTrickledownImpactReward2023, shaoSpuriousRewardsRethinking2025}, or conversely, sensitive to features \textit{that are not} causes (i.e., spurious correlates) \cite{liuRRMRobustReward2024, xuRewardModelsIdentify2025}.

Furthermore, when RMs are replaced with simpler metrics that explicitly reward surface-level features and not human preference, this often leads to minimal performance degradation in RLHF finetuning \cite{singhalLongWayGo2024, changBLEUBERIBLEUSurprisingly2025, gehrmannRewardModelsAre2025}. This can even be achieved with noisy or random rewards \cite{lvClimbCarvesWisdom2025, wangCausalRMCausalTheoreticReward2026}.

\parheading{Recovering RM Features} Cognizant of these findings, a recent line of research attempts to recover the features learned by RMs after training. Methods include regressing against a set of LLM annotated features \cite{liDissectingHumanLLM2024, revelSEALSystematicError2025}, training Sparse Auto Encoders (SAEs) on response embedding differences \cite{movvaWhatsMyHuman2025}, reinforcement learning prompt prefixes \cite{alazrakiReverseEngineeringHuman2025} or training an evolutionary LLM pipeline \cite{wangAutomaticallyFindingReward2026}. However, the relationship of these features to RM memorization has not yet been explored.

\begin{figure}[t]
    \centering
    \includegraphics{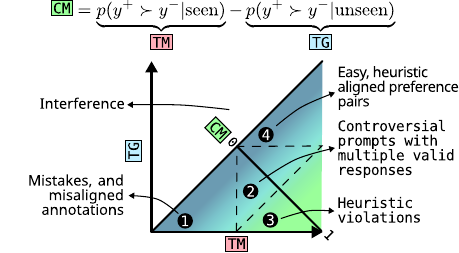}
    \caption{\textbf{Conceptual overview}: We measure counteractual memorization by taking the difference in probability of preference for an RM when a preference pair is or is not in the training data. We use this to construct a `memorization map', and try to determine what features are associated with different regions in this map.}
    \label{fig:hook}
\end{figure}

\parheading{Counterfactual Memorization \& RMs} Counterfactual memorization, as introduced by \citet{feldmanDoesLearningRequire2020} and \citet{zhangCounterfactualMemorizationNeural2023} provides a graded definition of instance-level memorization. \citet{dankersMemorisationCartographyMapping2023} use this to produce a memorization map (see Figure \ref{fig:hook}, allowing them to analyze which instance-level features impact memorization of machine translation samples.

While memorization in RLHF finetuned LLMs has been studied, this is primarily motivated by copyright or privacy concerns in LLM generations \cite{hartmannSoKMemorizationGeneralPurpose2023, ghoshRethinkingMemorizationMeasures2025}, to the best of our knowledge, no one has studied memorization in RMs specifically.

Both \citet{singhalLongWayGo2024} and \citet{leeDatasetCartographyLarge2025} produce dataset maps of human preference datasets, but using a distinctly different technique, and only for the purpose of estimating model confidence and data quality. These do not measure memorization, counterfactual or otherwise.


\begin{figure*}[h]
    \centering
    \includegraphics{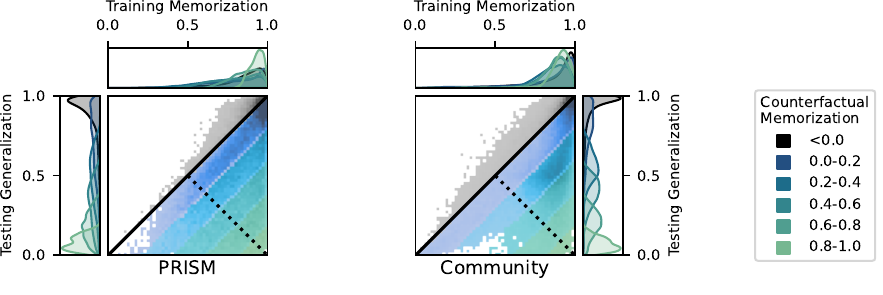}
    \caption{\textbf{Memorization maps}: \prism in the left pane and \community in the right. The \tm score is plotted along the horizontal axis, \tg along the vertical axis. The joint histogram is plotted centrally. For each bin, color denotes the \cm bin (see legend), and darkness the density of data present. The top and side panels give a KDE estimate of the marginal distributions of the preference pairs, for each \cm bin independently.}
    \label{fig:cm_maps}
\end{figure*}


\section{Quantifying Memorization}

Let $(x_{i},y_{i,1},y_{i,2},\ldots)\sim\mathcal{D}$ be a human preference dataset consisting of a user prompt $x$ (with conversational history prefixed), and a collection of candidate model responses $y$, ranked by humans for preference: $j<k\implies y_{i,j}\succ y_{i,k}$. We only consider the case of pairwise human preference feedback without ties, such that there is always one preferred ($y^{+}$) and one dispreferred ($y^{-}$) response for any preference pair.

\parheading{Datasets} We use two separate but related datasets: 1) \prism \cite{kirkPRISMAlignmentDataset2024}, and 2) \community \cite{zhangCultivatingPluralismAlgorithmic2025}. Both datasets include multi-turn conversations between users and chat models, with turn-level user preference labels provided by the same users. Both datasets provide substantive metadata at the user (\num{77}/\num{28}), conversation (\num{10}/\num{0}), turn (\num{1}/\num{0}) and response (\num{3}/\num{0}) levels (\prism/\community). 


\prism consists of \num{1.40e+3} users in \num{8.00e+3} conversations across \num{27.2e+3} turns. Turn \num{1} receives on average \num{3.75} responses from \num{21} different models, and later turns receive \num{2} responses from the same model. The balanced, English-only subset of \community consists of \num{0.77e+3} users in \num{17.8e+3} conversations across \num{45.8e+3} turns. Each turn receives \num{4} responses from the same model. In total, \prism has \num{57.9e+3} preference pairs, and \community has \num{137e+3}. We provide more detail in Appendix \ref{app:data}.

\parheading{Reward Modelling} An RM, symbolically denoted as $\texttt{rm}(y|x)$, is an approximation of the unobservable human preference function $r(y|x)$. It outputs a scalar `reward' for any $(x, y)$ pair. Discriminative RMs are trained to maximize the likelihood under some probabilistic model, most often the Bradley-Terry model \cite{bradleyRankAnalysisIncomplete1952}:
\begin{equation}
    p(y^{+}\succ y^{-})=\sigma(\mathtt{rm}(y^{+}|x)-\mathtt{rm}(y^{-}|x))
\end{equation}

We train RMs by finetuning the same backbone (\href{https://huggingface.co/meta-llama/Llama-3.2-1B}{\texttt{Llama-3.2-1B}} \cite{customLlama3Herd2024}, see Appendix \ref{app:model}) using a standard LM $\rightarrow$ SFT $\rightarrow$ RM pipeline \cite{zieglerFineTuningLanguageModels2019, wangSecretsRLHFLarge2024} common to training discriminative reward models \cite{zhongComprehensiveSurveyReward2025}. We use LoRA  \cite{huLoRALowrankAdaptation2021} to enable RM training within our budget. The full training methodology, along with a discussion on the interaction of PEFT with memorization, is provided in Appendix \ref{app:training}.

\parheading{Memorization Metrics} Following \citet{zhangCounterfactualMemorizationNeural2023} and \citet{dankersMemorisationCartographyMapping2023}, we use an operationalization of \textbf{counterfactual memorization} (\cm) as our measure of preference pair-level memorization. Specifically, we compute \cm by taking the difference in the expected probability of correct classification when a preference pair \textit{is} in the training data (which captures training memorization, \tm) and when it \textit{is not} (which captures testing generalization, \tg):

\begin{align}
    \begin{split}\cm&=\underbrace{\mathbb{E}[p(y^{+}\succ y^{-})|y^{+}, y^{-}\in\mathcal{D}^{\text{(train)}}]}_{\text{\tm}}\\&-~\underbrace{\mathbb{E}[p(y^{+}\succ y^{-})|y^{+}, y^{-}\not\in\mathcal{D}^{\text{(train)}}]}_{\text{\tg}}\end{split}
\end{align}

To estimate the \tm and \tg expectations, we train the same RM on \num{25} randomly sampled subsets of the same data. Each preference pair is present in only \num{20} of these subsets (\tm), and is held-out otherwise (\tg). To avoid user preference leakage, we perform this splitting at the user-level. We detail the full splitting procedure, including mitigating user-user confounding, in Appendix \ref{app:splitting}. A conceptual overview of counterfactual memorization can be found in Figure \ref{fig:hook}.

As we define it, \cm  is a purely relative metric; cases where both train and test performance are high are indistinguishable from cases where both are low. As such, we analyze all metrics separately, allowing us to distinguish between these scenarios.

\parheading{Model Scale} The generalization and memorization capabilities of LLMs both increase with model scale \cite{carliniQuantifyingMemorizationNeural2022, mahdaviMemorizationCapacityMultihead2024, morrisHowMuchCan2026}. To assess the robustness of our results, we re-run a reduced version of the experimental setup with a larger model backbone from the same family. Details are provided in Appendix \ref{app:model_scale}. We find that all memorization metrics are strongly correlated across model scale, and that these yield the same conclusions.

\parheading{Memorization Maps} We present the generated counterfactual memorization metrics in Figure \ref{fig:cm_maps}. For each preference pair, we treat its mean \tm (horizontal) and \tg (vertical) scores as a coordinate. Unsurprisingly, the vast majority of preference pairs are more often correctly classified after being included in the training data, and thus fall below the solid diagonal line. The dashed perpendicular line shows the direction of increasing \cm, which is maximal in the bottom right corner (coordinate $(1, 0)$).

After training, most of the preference pairs have already been memorized, resulting in most preference pairs falling on the right hand side of the map. Furthermore, most preference pairs have high \tg scores, and preference pairs with high \cm are rare, resulting in a dense cluster of preference pairs near coordinate $(1, 1)$. The few negative \cm preference pairs also tend to be near this point, which we attribute to noise.


\begin{figure*}[t]
    \centering
    \begin{subfigure}{0.49\textwidth}
        \includegraphics[width=\linewidth]{
            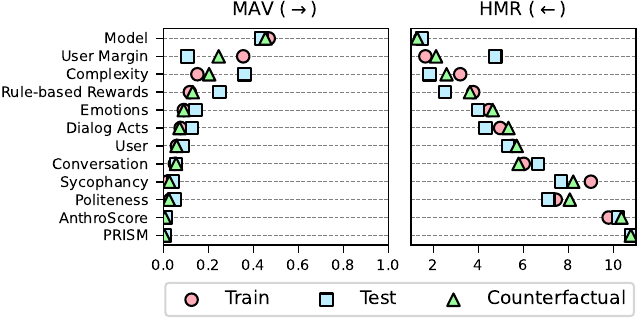
        }
        \caption{\prism}
        \label{fig:regression_prism_block_summary}
    \end{subfigure}
    \hfill
    \begin{subfigure}{0.49\textwidth}
        \includegraphics[width=\linewidth]{
            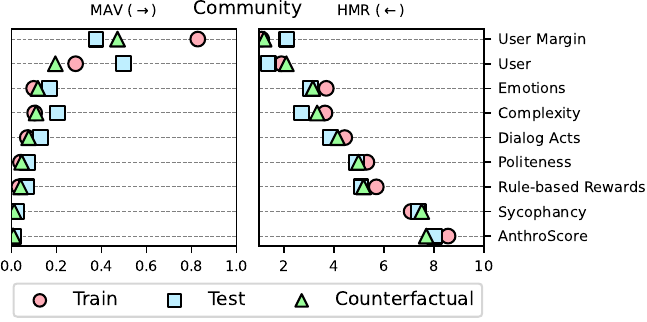
        }
        \caption{\community}
        \label{fig:regression_community_block_summary}
    \end{subfigure}
    \caption{\textbf{Block-level feature importance:} block-level net SHAP values, summarized by Mean Absolute Value (MAV) and Harmonic Mean Rank (HMR). Arrows point in direction of greater partial effect of the feature block on RM performance. The effect on \tm, \tg and \cm are denoted by a red circle, a blue square, and a green triangle, respectively. The more extreme a value, the more important that block as a whole is to RM performance. We order features by the MAV for \cm.}
    \label{fig:regression_block_summary}
\end{figure*}

\section{Hypothesis-Driven Determinants}
\label{sec:top_down}

With these dataset maps, we wish to study the determinants of \tm, \tg or \cm. What salient response properties make it easy to generalize to a response pair, and what properties require memorization?

\subsection{Feature Annotation}

We construct a set of features measuring various illocutionary speech acts within the context of user-LLM conversations, which we believe to be important correlates with either human or RM preference. In total, we provide \num{84} features per response, divided into the following blocks: \textbf{anthropomorphism} (\num{1}), measuring to which extent the LLM considers itself `human'; \textbf{complexity} (\num{9}), measuring the length, diversity, and complexity of a response; \textbf{dialog acts} (\num{23}), indicating whether a specific dialog act is present; \textbf{emotion} (\num{28}), indicating dominant emotions and sentiment; \textbf{politeness} (\num{4}), measuring how (im)polite a certain response is; \textbf{rule-based rewards} (\num{16}), indicating whether particular LLM behaviors are present; and, \textbf{sycophancy} (\num{3}), indicating to what extent the LLM support user framing. We use a combination of task-specific and general-purpose LLMs for automated annotation. We additionally supplement these features with the metadata provided by the dataset authors. More detail can be found in (see Appendix \ref{app:annotations}).

When organized into preference pairs, each feature represents a difference in that feature value between the chosen and rejected response. By convention, the more positive the difference, the more prevalent said feature is in the chosen response relative to rejected.

\subsection{Regression}

We treat each \cm component as an endogenous variable and regress against the feature differences. Specifically, we train a small \num{3} layer MLP regression model with the collected feature set as input and one of the endogenous variables as target output. The regressor is evaluated on a held-out evaluation set consisting of \qty{20}{\percent} of all preference pairs. To assess the importance of each feature to the performance of an RM, i.e., feature attribution, we use SHapley Additive exPlanations (SHAP) \cite{lundbergUnifiedApproachInterpreting2017} to estimate conditional Shapley values for each exogenous feature. We reserve \num{1e+3} preference pairs for perturbation and evaluate with all remaining samples.

\subsection{SHAP Summary Metrics} After the regression, we are left with an $N\times D$ matrix of SHAP values, which indicate to what extent a feature $d\in D$ contributed to the regression model's output of row $n\in N$. These are local explanations of model performance. To convert these to global, dataset-level explanations, we introduce several summary metrics. We describe these in detail in Appendix \ref{app:regression}. For the purposes of this section, we primarily focus on \num{2} summary metrics at the feature block level:
\begin{enumerate}[noitemsep, topsep=0.2em, labelindent = 0.0em, leftmargin = *]
    \item \textbf{Mean Absolute Value}: the average magnitude of the block net SHAP value
    \item \textbf{Harmonic Mean Rank}: the harmonic average of the block net SHAP value rank across a row (i.e., the inverse of the Mean Reciprocal Rank (MRR))
\end{enumerate}
MAV captures feature importance absolutely (higher being more important), whereas HMR captures this relatively to other active features for the same response pair (lower being more important).

We graphically depict these Figure \ref{fig:regression_block_summary}, based on Tables \ref{tab:regression_prism_shap_block} \& \ref{tab:regression_community_shap_block} in Appendix \ref{app:regression}. Summary statistics at the feature-level are provided in Appendix \ref{app:regression_individual} Tables \ref{tab:regression_shap_individual_values} and \ref{tab:regression_community_individual_values}.

Overall, the regression models predict RM performance reasonably well, achieving $R^{2}$ scores of \num{0.56}/\num{0.63}, \num{0.52}/\num{0.38} and \num{0.39}/\num{0.33} when regressing on \tm, \tg, and \cm, respectively (\prism/\community)\footnote{Features which are equally important to training or testing have no explanatory power on counterfactual memorization, which explains the reduced $R^{2}$-score.}. This indicates that RM performance is largely, and in some cases mostly, explained by a relatively small set of features.

Large differences in SHAP values for the same feature when used for different endogenous variables indicate that that feature is used differently for memorization or generalization. Specifically, large positive \cm features indicate \textbf{memorization}, being predictive of RM behavior after being seen during training, whereas large negative \cm features indicate \textbf{overgeneralization}, being less predictive of RM behavior when used during training.

\begin{figure}[t]
    \centering
    \includegraphics[width=\linewidth]{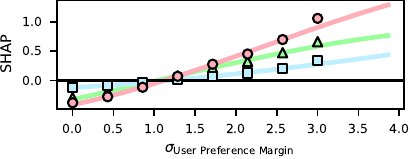}
    \caption{\textbf{User Margin partial effect}: mean SHAP value as a function of the (normalized) user margin for \prism. As the user margin increases, \tm increases quickly, but \tg remains relatively unaffected.}
    \label{fig:regression_prism_user_preference_margin}
\end{figure}

\parheading{User Margin} Features belonging to this block measure the the difference in user provided preference scores of the chosen and rejected response. For \prism, this is provided directly by the user, and is used in the reward model objective (Appendix \ref{app:training}). For \community, we impute this value from response annotations provided by different users on the same response, and is thus only available to the RM implicitly and across separate samples. 

In both datasets, user margin is a highly dominant predictor of \tm, but a substantially worse predictor of \tg, resulting in high \cm. Looking at SHAP correlation metrics in Appendix \ref{app:regression_individual}, the relationship between between \tm and preference margin is highly positive; the greater the margin, the better model performance is when the preference pair is included in the training data. This is clearly seen in Figure \ref{fig:regression_prism_user_preference_margin}; where \tm increases steadily with user margin, \tg stays close to \num{0.0} except for extreme values.

Thus, the model seems to recognize easy examples in its training data and maximizes the RM margin on these examples, but it cannot recognize these cases when held out. In other words, it appears that the model is memorizing cases where it already performs well but can safely maximize the RM score margin, thereby minimizing the dataset-wide Bradley-Terry loss. 

An ideal RM would operate in the opposite manner, memorizing difficult, controversial comparisons and generalizing to obvious ones. We deem this a misallocation of memorization capacity, which we discuss more in Section \ref{sec:discussion}).


\parheading{Model Identity} In \prism, responses can come from $21$ different models (see Appendix \ref{app:data} Figure \ref{fig:data_structure}). This metadata is only available post-hoc, although implicitly available to the RM through response model-specific style and behaviors.

In \prism it is the strongest determinant, implying that the RM is sensitive to these idiosyncratic differences in responses.

\begin{figure}[t]
    \centering
    \includegraphics[width=\linewidth]{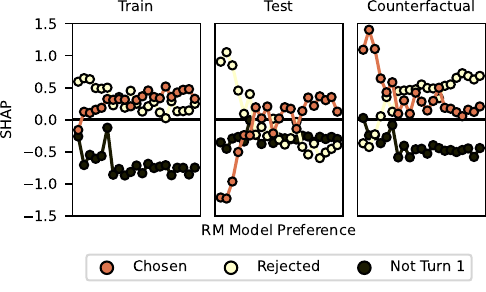}
    \caption{\textbf{Model identity partial effect}: mean SHAP value induced whenever a particular model is chosen (orange), rejected (tan) or is used for both chosen and rejected responses (black). Each pane gives the SHAP value for a different endogenous variable. Models are sorted by average RM score, with left-most being the `worst' performing model and right-most the `best'.}
    \label{fig:regression_prism_model_identity}
\end{figure}

We plot the SHAP values whenever a model is chosen or rejected in Figure \ref{fig:regression_prism_model_identity}, sorted by model-wise average RM score. The RMs have a strong preference for particular models---a preference that is positively correlated with human preference; see Appendix \ref{app:regression} Figure \ref{fig:regression_prism_user_rm_preference_correlation}---and high \tg SHAP values indicate that predictions on unseen preference pairs strongly correlates with textual style artifacts in the LLMs' generations. Specifically, when preference pairs are seen during training, performance degradation is moderate whenever a `better' model is rejected (or vice versa); however, when unseen, this degradation is substantial, especially for the least preferred models. This results in high \cm. When model differences are not present (the same model is used for all candidate responses), especially training performance decreases substantially.

Altogether, this suggests that the RMs associate idiosyncratic response models' output with user preference, and memorizes cases where this preference is violated.

\parheading{User} Whereas user-level features are only modestly predictive in \prism, it is the block with the 2nd highest \cm MAV in \community. This is predominantly due to a large reduction in RM performance on unseen preference pairs produced by users from the `(At most) Complete Secondary' education group. While recent work does suggest that users from different socio-economic backgrounds interact differently with LLMs \cite{bassignanaAIGapHow2025}, we do not find a difference between this group and other users in either English proficiency, or the distribution of discussed topics. Rather, the most significant difference seems to be in how these users were collected by \citeauthor{zhangCultivatingPluralismAlgorithmic2025}, with users from this group coming almost entirely from a second recruitment wave (see Appendix \ref{app:regression} Table \ref{tab:data_community_educ_wave}) specifically designed to increase the proportion of less educated individuals. The high \cm produced by preference pairs from these individuals indicate that the RMs are memorizing this user trait specifically, mediated through their prompts and preferences.

\begin{figure*}[t]
    \centering
    \begin{subfigure}{0.49\textwidth}
        \includegraphics[width=\linewidth]{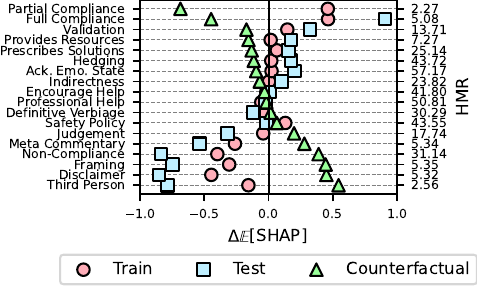}
        \caption{\prism}
        \label{fig:regression_community_block_summary}
    \end{subfigure}
    \begin{subfigure}{0.49\textwidth}
        \includegraphics[width=\linewidth]{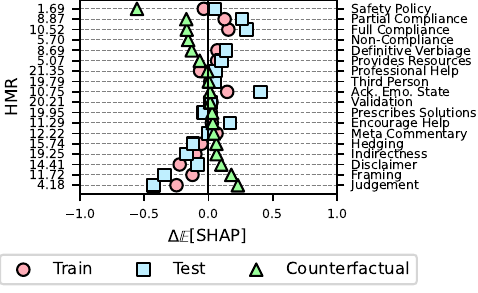}
        \caption{\community}
        \label{fig:regression_community_block_summary}
    \end{subfigure}
    \caption{\textbf{Rule-based rewards \& sycophancy partial effect}: the change in expected SHAP when there is a rule-based rewards or sycophancy feature difference across preference pairs: $\mathbb{E}[\text{SHAP}_{i,j}|\mathtt{feat}_{j}(y^{+}_{i})=1, \mathtt{feat}_{j}(y^{-}_{i})=0] - \mathbb{E}[\text{SHAP}_{i,j}|\mathtt{feat}_{j}(y^{+}_{i})=0, \mathtt{feat}_{j}(y^{-}_{i})=1]$. We interpret this as the effect of flipping a feature's presence in the chosen and rejected responses. The secondary axis provides the HMR whenever a feature difference is present (either positive or negative).}
    \label{fig:regression_rbr_partial_effect}
\end{figure*}

\parheading{Rule-Based Rewards \& Sycophancy} In Figure \ref{fig:regression_rbr_partial_effect} we graphically present differences in mean SHAP values for rule-based reward and sycophancy features when the feature is only present in the chosen response and only present in the rejected response, i.e., the effect of switching the chosen and rejected responses, all else held equal.

These features are especially predictive of memorization behaviors in \prism, as evidenced by the larger average deviations from the vertical null effect line. However, for both datasets, most rule-based rewards and sycophancy yield little difference in expected SHAP.

Partial or full compliance results in the largest gains in expected SHAP, for both \tm and \tg, indicating that the presence of this feature difference makes assigning a reward substantially easier. Conversely, the presence of refusal (`non-compliance') results in substantially worse performance. Additionally, at least in \prism\footnote{In \community, differences between responses are exceedingly rare, making it difficult to estimate reliable SHAP values.}, other variables indicative of non-compliance (e.g., `Disclaimer', `Meta-Commentary'), result in large gains in \cm. This indicates that the RMs naturally generalize to complying responses but must memorize refusals.

The sycophancy variables collectively contribute minimally to either training or testing performance. However, `Framing'---unquestioningly adopting a user frame---is individually associated with high \cm in both datasets due to a marked reduction in test-time performance when present in only the chosen response.

\parheading{Heuristics} The majority of our annotations result in larger MAV scores for \tg than \tm, increasing \cm. This implies that the RM depends on these features for unseen preference pairs, but attenuates their impact for seen preference pairs, i.e. overgeneralization.

Especially indicative of this behavior is the well-researched overreliance on differences in length, with a more verbose or complex response being more preferred by RMs, even in cases where this would be inappropriate. As a result, whenever the chosen response is more complex than the rejected response, this nets large negative \cm SHAP values, but when the chosen response is less complex, the opposite occurs (see Appendix \ref{app:regression_individual} Figure \ref{fig:regression_complexity_partial_effect}. In other words, the RMs naturally generalize to cases where the difference in the response pair aligns with the learned heuristic but must memorize cases where the learned heuristic is violated.

\begin{figure*}[t]
    \centering
    \includegraphics{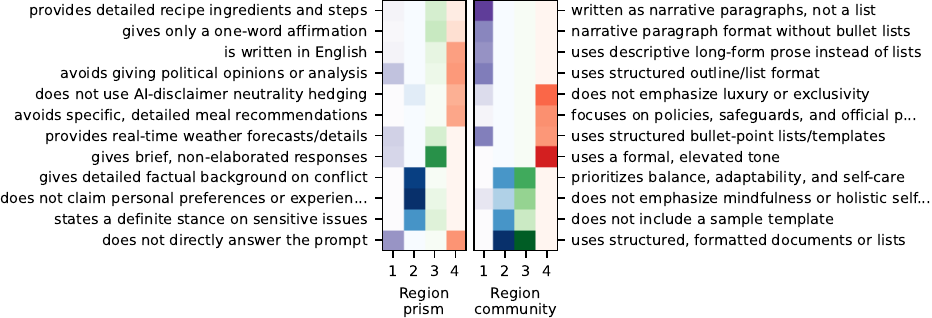}
    \caption{\textbf{SAE features for different \cm regions}: the feature descriptions of the conditional SAE latent space with the highest association weights with different regions in the counterfactual memorization map. These regions correspond to low/low (purple); medium/medium (blue); high/low (green); high/high (red) \tm/\tg, respectively. The darker the color, the greater the association between the SAE feature and that region.}
    \label{fig:sae_features}
\end{figure*}

\section{Data-Driven Determinants}
\label{sec:bottom_up}

To generate a set of data-driven features, we reimplement the reward model Sparse Autoencoders (SAE) of \citet{movvaWhatsMyHuman2025}. An SAE learns a set of spare latent variables that optimally reconstruct its input, and are often used in interpretability research \cite{shuSurveySparseAutoencoders}.

We first divide the memorization map into \num{4} disjoint regions (see Figure \ref{fig:hook}) before training a Matryoshka Batch Top-k \cite{bussmannBatchTopKSparseAutoencoders2024, bussmannLearningMultiLevelFeatures2025} SAE on the difference of the embeddings of the responses in a preference pair ($f(x, y^{+})-f(x, y^{-})$, where $f$ is the LLM). We allow for a maximum of \num{32} latent variables, of which an average of \num{4} are active for each sample. We use the same LM used to initialize the RMs to embed the responses. The SAE is then trained to minimize the $L2$ reconstruction loss, along with an auxilliary \cm region classification loss on the sparse latents. We interpret the weights of the region classifier (essentially a logistic regression) as associations between a region and a feature.

Following \citeauthor{movvaWhatsMyHuman2025}, we generate a textual description of each latent variable by feeding high scoring preference pairs to OpenAI \texttt{gpt-5}. These are then validated by annotating a subset of the data using OpenAI \texttt{gpt-5-mini} and checking the correlation against the SAE's active latent feature. We only retain the feature descriptions that maximally correlate with the active latent dimensions and which are significant at the \num{0.05} confidence level after Bonferroni correction.



The full list of discovered features can be found in Appendix \ref{app:sae} Table \ref{tab:sae_features}. We graphically display the features with the highest association weights in Figure \ref{fig:sae_features}.

For \prism there is clear differentiation in the discovered features between regions. The high \tm and high \tg region contains long, neutral responses that comply maximally with the user prompt. Contrasting this is the maximum \cm region, with low test generalization but medium to high train memorization, which is characterized by chosen responses that are brief, sometimes containing a single word, thereby violating the length difference heuristic function. The region with both medium train- and test-time performance contains references to non-neutral responses to controversial, politically charged topics. These are specifically cases where multiple responses might be preferred, depending on the stance of the user.

In \community, it is substantially more difficult to find feature differentiation, likely because all responses come from the same model, and differences are therefore more subtle. Formatting is a recurring feature across regions, but references are especially prevalent in the low \tm and low \tg region. Another recurring theme appears to be the response topic, which is especially discriminative between the middle two regions and the other two regions.

Overall, these results tentatively suggest that conventional, heuristic aligned preference pairs occur in the high/high region, whereas heuristic violating ones fall in the high/low region (i.e., inducing high \cm), and cases where there is a low-entropy response distribution (i.e., user-user preference conflict) correspond to the medium/medium region (\tm/\tg).

\begin{figure}[t]
    \centering
    \includegraphics[width=0.5\linewidth]{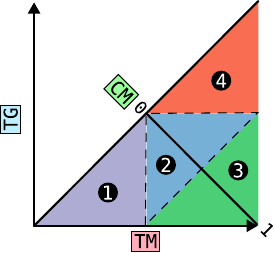}
    \caption{\textbf{SAE discriminative regions}: we divide the memorization map into \num{4} distinct, similar triangles. The colors and numbers match those used in Figure \ref{fig:sae_features}.}
    \label{fig:sae_discriminative_regions}
\end{figure}

\section{Discussion}
\label{sec:discussion}

For LLMs, memorization does not necessarily imply poor generalization and, in some cases, it is even beneficial to performance on unseen data. However, our results indicate that this is not happening; discriminative training of RMs on these datasets results in models that memorize dataset-specific non-causal correlates of human preference. We specifically identify \num{3} patterns in RM memorization:

{
\setlist{
    labelindent = 0.0em,
    leftmargin = *,
    topsep = 0.2em,
}
\begin{enumerate}
    \item \textbf{Misallocated memorization}: an ideal RM should supplement learned general rules with memorization of low-margin, difficult preference pairs. These include responses with subtle differences or on controversial topics. Instead, we see the opposite, with high margin preference pairs being more predictive of memorization than any other feature. These preference pairs are where it is safe to minimize training loss by maximizing the RM score margin without meaningfully increasing test-time performance.
    \item \textbf{Memorization of dataset artifacts}: besides user margin, the features that most clearly impact memorization deal with how the dataset was constructed: response model identity in \prism and user collection wave in \community. These features provide little information about human preference at large and will result in RMs that struggle to generalize outside their training data.
    \item \textbf{Overgeneralizing heuristic differences}: differences in simple to detect features that are correlates of human preference, like response length or compliance, are overgeneralized by the RMs. To effectively lower the training loss, violations of these heuristic differences are memorized, but for unseen response pairs further outside the training distribution, the default behavior reverts to rewarding these superficial differences, thereby enabling reward hacking.
\end{enumerate}
}

For memorization in RMs to complement generalization, these patterns will need to be addressed. This is especially important when  consistent, context-dependent judgments are necessary in settings far outside the RMs training distribution.

\section{Conclusion}
\label{sec:conclusion}

This paper studies what features RMs memorize during discriminative training on human-preference data. Our findings suggest \num{3} memorization patterns obstructing stronger, more robust RMs. These have direct implications for the design and evaluation of models of human preference. 

The prioritization of memorization of high margin preference pairs likely negatively impacts model robustness and might explain the recently reported effectiveness of curriculum learning in RM and RLHF \cite{pattnaikEnhancingAlignmentUsing2024, linCurriculumRLAIFCurriculumAlignment2026, liuMRACLMultiRewardSpace2026}. Analyzing the role of the Bradley-Terry objective on misallocated RM memorization should be a priority for the research community. Designing RM objectives that explicitly encourage complimentary memorization (i.e., of low margin preference pairs) may substantially boost performance in context-dependent cases.

Memorized dataset-specific shortcut features can inflate RM evaluation metrics and should be controlled for in the design of evaluation benchmarks. These should include data far removed from the RM training distribution, for example, by holding out particular user groups or response models. This was recently evidenced by \texttt{RewardBench 2} \cite{malikRewardBench2Advancing2025}, reporting a marked decrease in RM performance when filtering out prompts similar to those found in finetuning corpora.

Mitigating the overgeneralization of simple heuristics in RMs remains an open problem despite abundant attention. An especially promising direction is the inclusion of causally informed training, for example, by encouraging dependence on causally relevant features \cite{muRuleBasedRewards2024, wangInterpretablePreferencesMultiObjective2024}, or mitigating dependence on spurious correlates \cite{zhouExploreSpuriousCorrelations2024, yeRectifyingShortcutBehaviors2025, wangRewardHackingCausal2025, songCausalRewardAdjustment2026}, \textit{inter alia}.


Two direct extensions of this work are studying memorization behavior using alternative, non-discriminatively trained RMs (e.g., DPO \cite{rafailovDirectPreferenceOptimization2023}) or RMs trained on AI-generated feedback (i.e., RLAIF). The former might reveal how different algorithmic frameworks impact memorization, whereas the latter might show how pre-existing biases in LLMs can exarcerbate or mitigate dependence on dataset artifacts or shortcut features.

\section{Limitations}
\label{sec:limitations}

A key weakness of counterfactual memorization, as we have operationalized it in this paper, is the computational expense required in computing the separate training memorization and testing generalization terms. As a result, tradeoffs are required. First and foremost, we do not compute \cm at the preference pair-level but at the user-level. This reduces the total number of held-out permutation sets and simultaneously prevents user preference leakage, but runs the risk of conflating preference pair level differences with differences in users. However, the substantial within-user variance of \cm estimates dominates inter-user variance, suggesting that this is unlikely. Despite the drastic decrease in the number of held-out permutations, we remain constrained due to the prohibitive cost of finetuning LLMs into RMs. As a result, we are limited to only holding out each preference pair \num{5} times, resulting in potentially large variance in our estimates of \tg, and by extension \cm. This alone might already explain the reduced predictive power of the proxy regression models shown in Section \ref{sec:top_down}.

While we consider the results presented in Sections \ref{sec:top_down} and \ref{sec:bottom_up} reliable descriptors of RM performance, it should be noted that these are not causal estimates of the RMs inner mechanisms. Rather, they represent associations between a set of exogenous variables and the variables of interest. Establishing the true causal relation between data properties and RM memorization would require intervening on internal model representations using controlled edits.


\section{Ethical Considerations} 

We make extensive use of open-source and open-access research artifacts, and, in accordance with the licensing agreements, have done our best to adequately cite these and keep usage within the initial intended use case. We similarly intend to make our response-level annotations and memorization metrics publicly available to encourage further analysis into the memorization dynamics of RMs.

\section{Acknowledgements}

We would like to thank the reviewers for their helpful comments and feedback, which have significantly improved the quality of this work. We would also like to thank Vera Neplenbroek, Milan Miletić, Luan Fletcher, Ron Kremer, and Xiaoyu Tong for their insights on early drafts and the subsequent discussions.

This work was co-funded by the European Union (ERC, CulturAL, 101171968). Views and opinions expressed are however those of the author(s) only and do not necessarily reflect those of the European Union or the European Research Council. Neither the European Union nor the granting authority can be held responsible for them. \\[0.1em]

\par\noindent\includegraphics[height=4.0em]{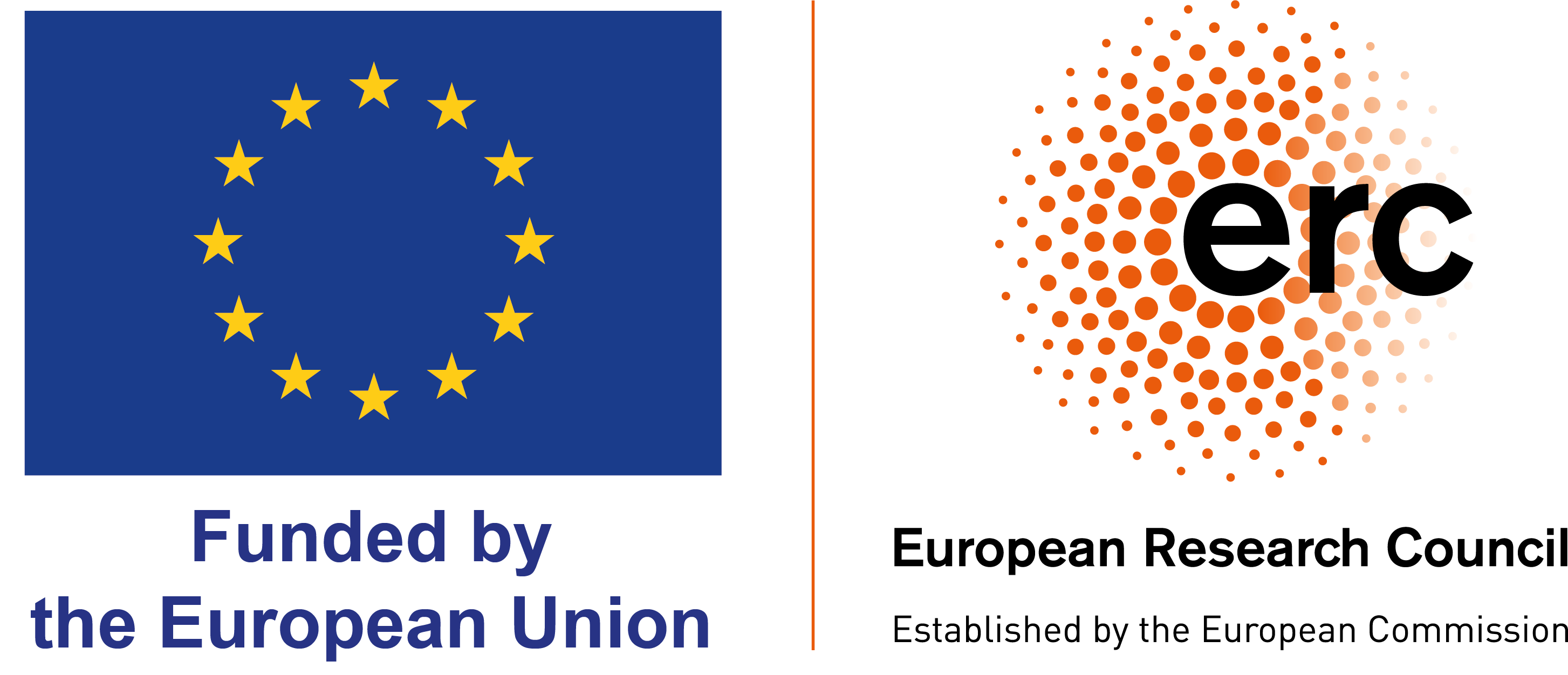} \\[-2.5em]


\bibliography{bib/zotero,bib/custom}


\clearpage

\appendix

\begin{figure*}[!ht]
    \centering
    \includegraphics[width=\textwidth]{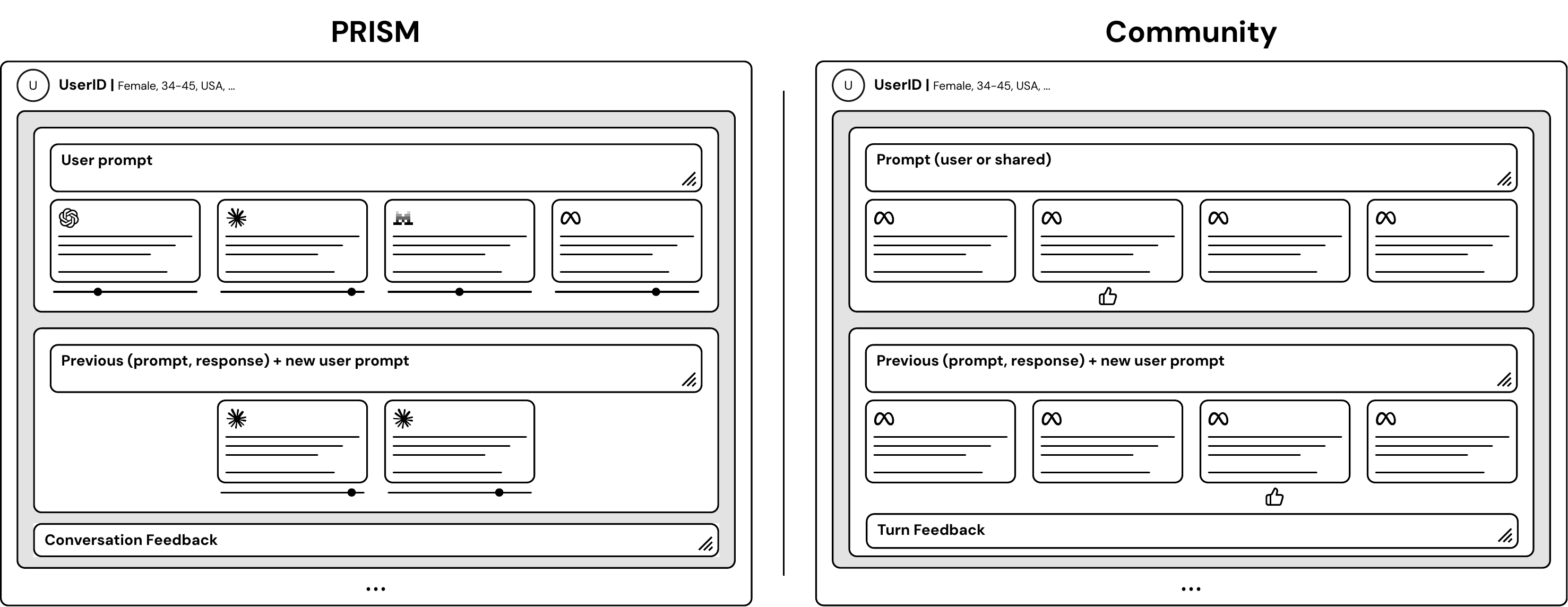}
    \caption{\textbf{Dataset structure}: Left is \prism, right is \community. Both datasets provide user-level metadata, and have the user interacting with the LLM also rate each response. In \prism, these responses come from \num{1} of \num{21} models in the first turn, and the user rates each response on an ordinal scale, with the LLM providing the best response being used for subsequent turns. In \community, the user selects the best response from \num{1} of \num{4} response from the same LLM. Sometimes the user is provided with the starting prompt, and sometimes the user leaves turn-level, free text motivations for their choice.}
    \label{fig:data_structure}
\end{figure*}

\etocdepthtag.toc{mtappendix} \etocsettagdepth{mtchapter}{none} \etocsettagdepth{mtappendix}{section}
\tableofcontents


\section{Data}
\label{app:data}

We use two separate, but similarly structured, human preference datasets: the 1) \prism \cite{kirkPRISMAlignmentDataset2024}, and 2) \community \cite{zhangCultivatingPluralismAlgorithmic2025} alignment datasets. Both datasets include user-LLM conversations with preference labels produced by the same user providing the prompts; both provide extensive metadata on said users; and both aim to maximize the socio-cultural diversity in their user pool to incorporate heterogeneous preferences.

Each user produces many conversations, which typically cover a single topic. In \prism, users free to discuss any topic but are prompted to provide the type of conversation they intend to have (unguided/controversy guided/values guided). In \community the users are usually free to determine the conversation topic, but sometimes ($\approx2\%$) users are presented with a pre-generated initial prompt. This ensures prompt overlap between users, which is useful for studying pluralistic conflict.

Each conversation contains at least one turn of interaction, which consists of a prompt, a set of LLM-generated candidate responses, and a user-generated preference label for each candidate response. In \prism, preference is annotated using an ordinal $1-100$ scale (least preferred to most preferred), whereas in \community, users only annotate which of the $4$ responses they preferred most. The former allows for estimating a margin of preference, and we can train preference on all combinations of responses, whereas in the latter case, it is impossible to determine the order of preference between dispreferred responses.

The LLM canidate responses come from \num{1} of \num{21} models different models in the first turn of conversations in \prism. The `winning' model is then used to source all candidate responses in later turns. The \community authors take a different approach, using the same \verb|Llama-3.3-70B-Instruct| to generate all model responses, although sampled in such a manner as to increase diversity in response stances. 

Besides preference labels, users sometimes provide natural language feedback. In \prism this occurs at the conversation level, whereas in \community this is at the preference pair-level.

We depict the described structure of the datasets in Figure \ref{fig:data_structure}.

\subsection{User Metadata and Diversity}

Both datasets were explicitly created for the purpose of evaluating and developing cultural sensitivity in large language models under a RLHF framework. However, the approaches used to cultivate diversity in user preference and model response differ substantially.

With \prism, \citeauthor{kirkPRISMAlignmentDataset2024} collect prompt-response pairs $1396$ from users in $38$ separate countries. Each user additionally provides information about their socio-cultural identity (e.g., gender, age, ethnicity), along with individual preferences (as measured using a 7 item ordinal scale) and their familiarity with LLMs.

To diversify the LLM responses provided to the users, in turn 1, the user is presented with up to $4$ responses produced by a pool of $21$ different models. In later turns, only the highest scoring LLM from turn 1 is used.

While this approach ensures diversity in user-level attributes and model identities, \prism assumes that a diverse set of users or LLMs results in a diverse set of user preferences or LLM responses. However, \citeauthor{zhangCultivatingPluralismAlgorithmic2025} show that this is not the case, with high homogeneity at the socio-cultural level between responses, despite seemingly high surface-level variation.

With \community, \citeauthor{zhangCultivatingPluralismAlgorithmic2025}
attempt to correct this. They use a similar setup, collecting user-level socio-cultural metadata from $3196$ users from $5$ separate countries. We select only users conversing in English, which reduces the pool of countries to India and the United States.

As a rough estimate for the amount of socio-cultural variation within each dataset, we provide a cross tabulation of the proportion of respondents in each ethnicity-location/country pairing in Tables
\ref{tab:data_cultural_diversity_prism} and \ref{tab:data_cultural_diversity_community}. Note that \prism has a substantially larger  pool of locations, and that both datasets have a similar dominant group (White-USA in \community, and White-USA/UK in \prism). Despite the reduced variance in socio-cultural groups in \community, \citeauthor{zhangCultivatingPluralismAlgorithmic2025} claim a much higher degree of socio-cultural variation in LLM responses.

\begin{table*}
    [t]
    \begin{tblr}{
    width = 1.0\textwidth,
    colspec = {lX[-1, c]X[-1, c]X[-1, c]X[-1, c]X[-1, c]X[-1, c]X[-1, c]X[-1, c]X[-1, c]X[-1, c]X[-1, c]X[-1, c]},
    hline{1,3,Y,Z} = {solid},
    vline{Y} = {1-Z}{solid, abovepos = -1},
    row{1} = {font=\bfseries\small},
    row{2} = {font=\itshape\small},
    cell{1}{1} = {r=2,c=1}{c},
    cell{1}{2} = {r=2,c=1}{c},
    cell{1}{3} = {r=1,c=2}{c},
    cell{1}{5} = {r=2,c=1}{c},
    cell{1}{6} = {r=1,c=2}{c},
    cell{1}{8} = {r=1,c=4}{c},
    cell{1}{12} = {r=2,c=1}{c},
    cell{1}{13} = {r=2,c=1}{c},
    row{3-14} = {font=\small},
    abovesep = -2pt,%
    belowsep = -2pt
}
\diagbox[width=5em]{Eth}{Loc}
& Africa      & Asia &      & Aus. \& NZ & Americas &       & Europe & & &                      & PntS & Total \\
& & East & West &            & North    & Latin & North & South & East & West &                   &       \\
Asian & 0.13 & 1.20 & 0.07 & 1.40 & 2.33 &  & 1.13 &  &  & 0.07 &  & 6.33 \\
Black & 4.80 &  &  & 0.13 & 2.27 &  & 0.93 &  &  &  &  & 8.13 \\
Hispanic &  &  &  &  & 1.67 & 6.27 & 0.07 & 0.07 &  &  &  & 8.07 \\
Mixed & 0.33 &  & 0.20 & 0.67 & 1.47 & 0.60 & 0.87 & 0.13 &  & 0.27 &  & 4.53 \\
Other &  &  & 1.60 & 0.33 & 0.20 & 0.07 & 0.33 &  &  & 0.07 &  & 2.60 \\
White & 0.27 &  & 1.73 & 6.93 & 20.00 & 0.80 & 24.07 & 3.80 & 2.87 & 4.07 & 0.07 & 64.60 \\
PntS & 0.33 &  & 0.47 & 0.47 & 1.40 & 1.13 & 1.20 & 0.07 & 0.07 & 0.60 &  & 5.73 \\
Total & 5.87 & 1.20 & 4.07 & 9.93 & 29.33 & 8.87 & 28.60 & 4.07 & 2.93 & 5.07 & 0.07 & 100.00 \\
\end{tblr}
    \caption{Proportion of \prism respondents who self-report belonging to
    an Ethnicity-Location pair (columns and rows, respectively). Proportions
    are provided as percentages.}
    \label{tab:data_cultural_diversity_prism}
\end{table*}

\begin{table}[t]
    \begin{tblr}{
    width = \linewidth,
    colspec = {lX[-1, c]X[-1, c]X[-1, c]},
    hline{1,3,Y,Z} = {solid},
    vline{Y} = {1-Z}{solid, abovepos = -1},
    row{1} = {font=\bfseries\small},
    cell{1}{1} = {c=1,r=2}{c},
    cell{1}{2} = {c=1,r=2}{c},
    cell{1}{3} = {c=1,r=2}{c},
    cell{1}{4} = {c=1,r=2}{c},
    row{2-Z} = {font=\small},
    abovesep = -2pt,%
    belowsep = -2pt
}
\diagbox[width=5em, height=2em]{Eth}{Cntr}
& India & United States & Total \\
& & & \\
Asian &  & 2.94 & 2.94 \\
Black &  & 4.33 & 4.33 \\
Dravidian & 14.25 &  & 14.25 \\
Hispanic &  & 6.66 & 6.66 \\
Indo-Aryan & 39.10 &  & 39.10 \\
Other & 2.75 & 0.44 & 3.19 \\
White &  & 24.17 & 24.17 \\
PntS &  & 5.34 & 5.34 \\
Total & 56.11 & 43.89 & 100.00 \\
\end{tblr}
    \caption{Proportion of \community respondents who self-report belonging
    to an Ethnicity-Location pair (columns and rows, respectively).
    Proportions are provided as percentages.}
    \label{tab:data_cultural_diversity_community}
\end{table}

\section{Splits}
\label{app:splitting}

To estimate \cm, we generate \num{25} unique splits of the data and exclude each sample from \num{5} random splits. To ensure there is no data leakage of user preferences between the splits, we perform the exclusion at the user level. This means all user interactions occur in either $\mathcal{D}^{\text{(train)}}$ or in $\mathcal{D}^{\text{(test)}}$, but never both.

In an ideal experiment, we would produce \num{1} split per user. However, due to the prohibitive cost of RM finetuning, and the inherent stochasticity of deep learning with mini-batch gradient descent, we are forced to use significantly fewer splits.

A purely uniform assignment of users to hold out test splits will inevitably lead to high user-user co-occurrences across splits. While unavoidable with only \num{5} hold out splits per user, this does introduce an undesirable correlation into the \cm computation. To mitigate this, we initialize an assignment by greedily placing a user in the bucket with users it has seen the least. After this, we search for a better assignment that minimizes the overall user-user co-occurrences. We repeat this for \num{2e+3} iterations, keeping only the best solution. Overall, this ensures a user co-occurs with other users \textit{at most} \num{3}/\num{5} times, where both random assignment and the greedy solution see frequent \num{4}/\num{5} or \num{5}/\num{5} user-user co-occurences.

For all other experiments, whenever hold out splits are required, we simply use standard preference pair splitting.

\section{Model}
\label{app:model}

When estimating \cm, we finetune a \href{https://huggingface.co/meta-llama/Llama-3.2-1B}{\texttt{Llama-3.2-1B}} \cite{customLlama3Herd2024} baseline with LoRA \cite{huLoRALowrankAdaptation2021} adapters through the \href{https://huggingface.co/docs/peft}{\texttt{PEFT}} library \cite{mangrulkarPEFTStateoftheartParameterEfficient2022}. Specifically, we apply LoRA to all projection weights in the model, set $\mathtt{r}=32,~\alpha=32$ and initialize using EVA \cite{paischerParameterEfficientFinetuning2025}. This results in a total of \num{22.5d+6} parameters, representing just \qty{2}{\percent} of all available parameters.

The effect of PEFT on memorization has received some attention in prior works. Overall, the scientific consensus is relatively unanimous; at the same level of performance, a model fine-tuned with PEFT memorizes less of the training data than a full finetune. Specifically, \citet{mireshghallahEmpiricalAnalysisMemorization2022} find that models finetuned with adapters memorize less of the training data at the same level of validation perplexity. In general, \citet{bossyMitigatingUnintendedMemorization2025} find models trained with PEFT memorize substantially less, which \citet{hongEvaluatingMemorizationParameterefficient2025} corroborate and partially attribute to the PEFT parameters acting as an information bottleneck. This is evidenced by \citet{houImpactFineTuningMethods2025}, who find that small(er) parameter updates memorize less across different PEFT techniques. However, \citet{wangLeanerTrainingLower2025} find that PEFT finetuned models do still memorize, just that this is substantially less than full model finetuning. In general, 

As such, given the weight of prior evidence and the empirical findings in the next section, we (1) believe that our PEFT finetuned models are capable of memorizing the datasets and (2) expect that our findings generalize to fell model reward training setups.

\subsection{Model Scale}
\label{app:model_scale}

While \href{https://huggingface.co/meta-llama/Llama-3.2-1B} is a sufficiently competent model to serve as a backbone for RM finetuning, most SoTA RMs use substantially larger LLMs as backbones. Model scale tends to positively increase model competency, by increasing both model generalisation and memorisation \cite{carliniQuantifyingMemorizationNeural2022, mahdaviMemorizationCapacityMultihead2024, morrisHowMuchCan2026}. As such, model scale can influence the results of our experiments. 

To this directly, we re-run the experiment with a 3B parameter Llama-3.2 backbone instead. To accomodate our computation budget, we compute \tm, \tg and \cm estimates using 8 train + 2 test splits per sample. This produces estimates of each quantity in the exact manner, but the estimates have larger variance. As such, we do not draw conclusions from these estimates directly, but rather, use it to confirm the findings we present with the smaller model (i.e., a robustness check). We opted to stay within the same family and generation of models to isolate the effect of scale, rather than differing pre-training data or training setups. 

\begin{figure*}
    \centering
    \includegraphics[width=1.0\linewidth]{
        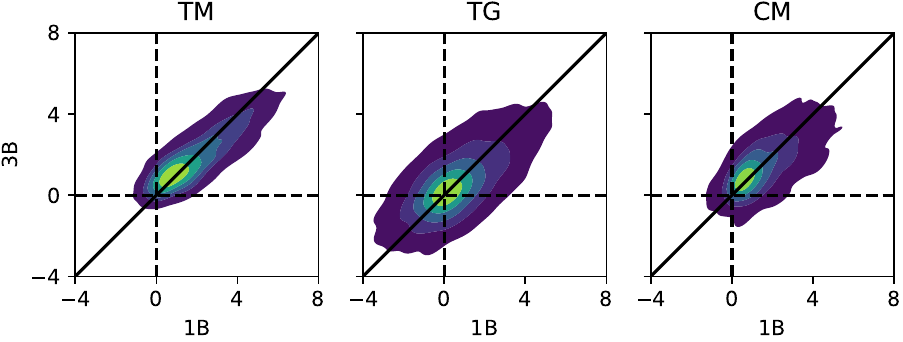
    }
    \caption{Correlation of \tm, \tg, and \cm estimates on \prism when using \href{https://huggingface.co/meta-llama/Llama-3.2-1B}{\texttt{Llama-3.2-1B}} (horizontal) or \href{https://huggingface.co/meta-llama/Llama-3.2-3B}{\texttt{Llama-3.2-3B}}  (vertical) as the backbone for the RMs.}
    \label{fig:1b_vs_3b_correlation}
\end{figure*}

\begin{figure}
    \centering
    \includegraphics[width=\linewidth]{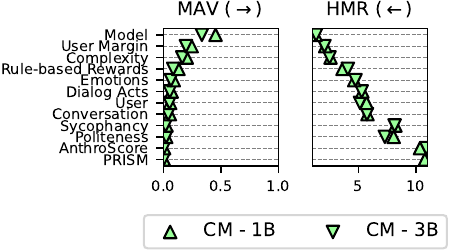}
    \caption{SHAP summary metrics produced by regression models of \cm using \href{https://huggingface.co/meta-llama/Llama-3.2-1B}{\texttt{Llama-3.2-1B}} (upwards triangle) or \href{https://huggingface.co/meta-llama/Llama-3.2-3B}{\texttt{Llama-3.2-3B}}  (downwards triangle) as the backbone for the RMs.}
    \label{fig:1b_vs_3b_regression_correlation}
\end{figure}

Overall, we find that all metrics' estimates are strongly correlated across model scale. This can be clearly seen in Figure \ref{fig:1b_vs_3b_correlation}. Plotted against each other, especially \tm is closely aligned with the 1-1 correspondence line (solid, diagonal line), although it tends to be slightly larger for the 3B model. This implies that the larger model is memorizing the data slightly more. \tg is less aligned, although still clearly correlated; performance on unseen data samples did not increase by much. The additional variance, relative to \tm, can be explained by the lower number of total training runs. Ultimately, this gives \cm, which is also strongly correlated across model scales, is also slightly larger for the 3B model, and has variance that lies between that of \tm and \tg.

We additionally repeat the regression analysis. At the feature block level, we find that the net coefficients are remarkably similar. This is clearly visible in Figure \ref{fig:1b_vs_3b_regression_correlation}. The MAV SHAP values are close, although the 3B parameter model consistently has lower values (which we again attribute to increase variance due to fewer training runs), and the relative importance of the feature blocks to the regression model overlaps nearly perfectly. For \prism, Model Identity, User Margin, Complexity, and---to a lesser extent---Rule-based Rewards are the most predictive of CM.

While we cannot use this analysis to draw conclusions about much larger models (i.e., >10B), we do believe it adds substantial credence to the robustness of our analysis. Different model scales evaluated on the same dataset should lead to the same conclusions.

\section{RM Training}
\label{app:training}

\begin{figure*}[t]
    \centering
    \includegraphics[width=1.0\linewidth]{
        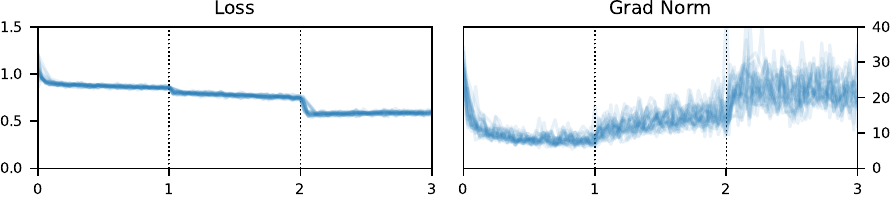
    }
    \caption{Loss and gradient norm curves for all 25 model iterations when
    training on \prism. Dotted black lines denote epoch boundaries.}
    \label{fig:prism_loss_short_training_runs}
    \includegraphics[width=1.0\linewidth]{
        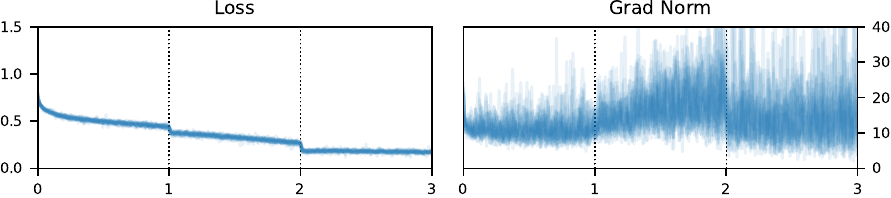
    }
    \caption{Loss and gradient norm curves for all 25 model iterations when
    training on \community. Dotted black lines denote epoch boundaries.}
    \label{fig:community_loss_short_training_runs}
    \includegraphics[width=1.0\linewidth]{
        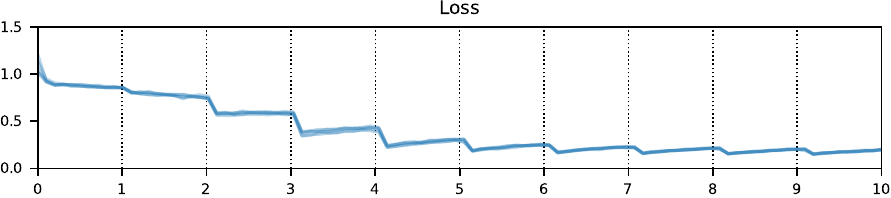
    }
    \caption{Loss and gradient norm curves for all 25 model iterations when
    training on \prism. Dotted black lines denote epoch boundaries.}
    \label{fig:prism_loss_long_training_runs}
\end{figure*}

We fine-tune the base LLM on \num{25} different splits of each dataset using \href{https://huggingface.co/docs/trl/index}{\texttt{TRL}}
\cite{vonwerraTRLTransformersReinforcement2020} (v19.1). We first apply a single epoch of supervised finetuning (SFT) on all chosen responses, conditioned on the prompt and conversation history. We apply linear warmup for the first \qty{2}{\percent} of steps, before using a constant learning rate. This SFT step ensures the model can use the chat format corectly.

This is followed by \num{3} epochs of Bradley-Terry reward model finetuning. The validation loss and accuracy were optimal after a single epoch of RM training, however we chose \num{3} epochs for computing the memorization metrics since it:
\begin{itemize}
    \item is $3$ times longer than necessary for convergence, matching
        \citet{dankersMemorisationCartographyMapping2023}, who explicitly note training beyond convergence in order to investigate memorisation (p. 8335)

    \item empirically, longer training runs saw substantially worse validation scores

    \item more epochs for all splits did not fit within our compute budget
\end{itemize}

Besides a pure Bradley-Terry classification loss, we also include a
centering loss as a regularizer on the predicted rewards \cite{eisensteinHelpingHerdingReward2024}. For the models trained on \prism, where users provide a scalar preference score for each response, we additionally include a margin:
\begin{equation*}
    \sigma(\mathtt{RM}(y^{+}|x)-\mathtt{RM}(y^{+}|x)+\underbrace{(r(y^{+}|x)-r(y^{-}|x))}
    _{\text{user margin}})
\end{equation*}
Rather than providing raw user margins, we bin these values using the 25th, 50th, 75th and 100th quantiles of margins, mapping these bins to scores of $\{0, 1/ 3, 2/3, 1\}$, respectively, matching \citet{touvronLlama2Open2023} (see Appendix A.3.3 Table 27). They observe that ``... the margin term can indeed help the reward model perform better on more separable comparison pairs ... [it] also regresses performance on similar samples.''

During RM training and CM computation, we remove all samples with token lengths greater than $1400$. We use the same learning-schedule as SFT, and optimise using \texttt{AdamW} \cite{kingmaAdamMethodStochastic2015,loshchilovDecoupledWeightDecay2019}.

We present the training loss and gradient norm curve for \prism in Figure \ref{fig:prism_loss_short_training_runs}. Overall, the training loss curves were vey consistent between runs, despite seeing different subsets of the same data. We observe a consistent sharp drop in loss after each epoch, with an accompanying sharp increase in average gradient norm for the first $3$ epochs of training. This pattern continues for much longer training runs\footnote{Due to the prohibitive cost of these training runs, only $5$ runs were continued for $10$ epochs} and indicates that the model has successfully memorized all data samples (see Figure \ref{fig:prism_loss_long_training_runs}).

When training RMs on \community we use the same methodology, except that we do include a user margin (\community does not provide scalar user preferences for responses). We present the loss and gradient curve for training on \community in Figure \ref{fig:community_loss_short_training_runs}.

Overall, the RMs achieve slightly lower accuracy scores on \prism (\qty{63}) compared to \community (\qty{69}{\percent}). In both instances, this is slightly lower than the optimal accuracy score achieved after a single epoch of training;\prism \qty{65} and \community \qty{72}{\percent}.

\section{Annotations}
\label{app:annotations}

\begin{table*}[t]
    \begin{tblr}{
    width = \linewidth,
    colspec = {X[-1, l]X[-1, c]ccX[-1, j]},
    hline{1,2,Z} = {solid},
    hline{3-10} = {solid, fg=lightgray},
    row{1} = {font=\bfseries\small},
    cell{2}{1} = {c=2,r=1}{l},
    cell{3}{1} = {c=1,r=3}{l},
    cell{6}{1} = {c=2,r=1}{l},
    cell{7}{1} = {c=2,r=1}{l},
    cell{8}{1} = {c=2,r=1}{l},
    cell{8}{1} = {c=2,r=1}{l},
    cell{8}{1} = {c=2,r=1}{l},
    row{2-Z} = {font=\small, valign=m},
}
    Block & Sub-Block & Domain & Number & Description \\

    Antrhopomorphism & & $\mathbb{R}$ & 1 & Estimates to which degree the response assigns animacy to itself using the \texttt{AnthroScore} framework \cite{chengAnthroScoreComputationalLinguistic2024}. Specifically, we mask all 1st person pronouns within the response, and then use a masked language model to determine the probability of an animate versus an inanimate pronoun replacement \\

    Complexity
    & Length & $\mathbb{R}$ & 3 & number of tokens, number of unique lemmas, number of sentences \\
    & Complexity & $\mathbb{R}$ & 5 & average sentence length, average number of syllables per token, number of sentences, number of entities \\
    & Diversity & $\mathbb{R}$ & 1 & type-to-token ratio \\

    Dialog Acts & & $\mathbb{R}$ & 23 & We get logits for \textsc{midas} \cite{yuMIDASDialogAct2021} dialog acts for the prompt and first sentence of the response using \href{https://huggingface.co/CLTL/midas-da-xlmroberta}{\texttt{CLTL/midas-da-xlmroberta}} \cite{LeolaniCltldialogueclassification2025} \\

    Emotion & & $\mathbb{R}$ & 28 & We get logits for
    \textsc{goemotions} \cite{demszkyGoEmotionsDatasetFineGrained2020} emotions over the entire response using \href{https://huggingface.co/AnasAlokla/multilingual_go_emotions_V1.2}{\texttt{AnasAlokla/multilingual\_go\_emotions}} \\

    Politeness & & $\mathbb{R}$ & 4 & We get logits for
    politeness levels over the entire response using \href{https://huggingface.co/Intel/polite-guard}{\texttt{Intel/polite-guard}} \cite{IntelPoliteguard2026}. This model was specifically trained on human-LLM interactions \\

    Rule-based Rewards & & \{0,1\} & 16 & We annotate responses for possessing one of the identified LLM behaviors in \citet{muRuleBasedRewards2024}. We use their prompts and annotate automatically using \href{https://developers.openai.com/api/docs/models/gpt-4.1-nano}{\texttt{OpenAI/gpt-4.1-nano}}. For comparisons, this feature is \num{+1} if it is present in chosen but not in the rejected response, \num{-1} if \textit{vice versa} and \num{0} otherwise \\

    Sycophancy & & \{0,1\} & 3 & We annotate responses for possessing one of the identified sycophancy traits in \citet{chengELEPHANTMeasuringUnderstanding2025}. We use their prompts and annotate automatically using \href{https://developers.openai.com/api/docs/models/gpt-4.1-nano}{\texttt{OpenAI/gpt-4.1-nano}}. For comparisons, this feature is \num{+1} if it is present in chosen but not in the rejected response, \num{-1} if \textit{vice versa} and \num{0} otherwise \\

\end{tblr}

    \caption{Manually annotated features.}
    \label{tab:top_down_features}
\end{table*}

Let $x_{i}$ be a user prompt (potentially with prepended conversational history), and $y_{i}^{+}$, $y_{i}^{-}$ be the corresponding user preferred and dispreferred responses. We annotate each preference pair like this with a set of features using some annotation method: $\mathrm{feat}_{j}^{\text{user}}(x_{i},y_{i}^{+}, y_{i}^{-})$

We selected \prism and \community primarily for the user-level metadata annotations. Specifically, \prism provides user information about demographic factors (e.g., age, gender, ethnicity, location), socio-economic status (e.g., education, employment, marital status), English \& LLM proficiency, along with self-reported descriptions and values. \community provides a smaller subset of user-level features, focusing on demographics and education, but also provides political stances. This information is constant across preference pairs and might allow us to detect inter-user preference conflict. We represent these as boolean dummy variables, $\mathrm{feat}_{j}^{\text{user}}(x_{i},y_{i}^{+}, y_{i}^{-})\in \{0,1\}$.

\prism additionally provides annotations at the conversation and response levels. Specifically, each conversation is rated along \num{10} axes for particular facets of quality, and each response has been annotated automatically for non-English texts, personally identifiable information (PII) and violations of a moderation policy. The former is constant across the preference pair, and is thus represented as a boolean dummy variable, $\mathrm{feat}_{j}^{\text{conv}}(x_{i},y_{i}^{+}, y_{i}^{-})\in \{0,1\}$, whereas for the latter we model differences using a ternary integer representation, $\mathrm{feat}_{j}^{\text{turn}}(x_{i},y_{i}^{+}, y_{i}^{-})\in \{-1,0,1\}$. Here \num{1} indicates the feature is present in the chosen response but not in the rejected response, \num{-1} indicates the opposite, and \num{0} indicates that the feature is either present in both chosen and rejected, or in neither.

Another response-level feature provided in \prism, is the user preference score, as measured on a \numrange{1}{100} ordinal scale. We represent the preference pair-level feature, user margin, as a continuous variable expressing the difference in preference score: $\mathrm{feat}_{j}^{\text{turn}}(x_{i},y_{i}^{+}, y_{i}^{-})\in \{1,99\}$. This feature is not present in \community, so we estimate a user margin on the preference pairs repeated across different conversations. A more detailed description is provided in Appendix \ref{app:regression_individual}. 

Finally, we complement these features by annotating responses using a combination of pretrained standard NLP toolkits, open-source models, LLM APIs. These features and their preference pair-level representation domain can be found in Table \ref{tab:top_down_features}.

Overall, we classify all the features we use into the following categories:

\begin{enumerate}
    \item \textbf{Dataset Native Annotations}: feature blocks \textit{User Margin}, \textit{Model Identity}, and \textit{User} all use metadata provided by the datasets we analyze. The datasets were specifically chosen for these metadata features
    \item \textbf{Low-Level NLP Tools}: for the \textit{Complexity} feature block, we use standard, low-level NLP tools like SpaCy and the RMs tokenizer. Both are general-purpose tools and have years of active use in research
    \item \textbf{Masked Language Modelling}: \textit{Anthropomorphism} uses the \texttt{AnthroScore} framework \cite{chengAnthroScoreComputationalLinguistic2024}, which relies on a general-purpose masked language model (MLM) to compute the probability of various animate and inanimate pronouns. We use a SoTA MLM (\texttt{ModernBERT} \cite{warnerSmarterBetterFaster2025}), and we replicate their methodology exactly otherwise
    \item \textbf{External Classifiers}: for the \textit{Dialog Acts}, \textit{Emotions}, \textit{Politeness} feature blocks, we use the logits of externally fine-tuned classification models as features. We do not softmax normalize these at the instance level, instead standardizing per feature at the dataset level. We assume that at the instance level, the relative magnitude of logits (i.e., classification performance) will suffer due to cross-dataset generalization. However, this should not affect the quality of our features; for example, we are not interested in a classifier that correctly identifies ‘anger’ as the dominant emotion, but rather one that increases its logits if an indicator of ‘anger’ is present
    \item \textbf{External LLMs}: for the \textit{Rule-based Rewards} and \textit{Sycophancy} feature blocks, we use the same external LLM (OpenAI/gpt-4.1-nano) with prompts provided by \citet{muRuleBasedRewards2024} and \cite{chengELEPHANTMeasuringUnderstanding2025}. These prompts were developed for specifically for analyzing LLM responses
\end{enumerate}

Thus feature blocks \textit{Anthropomorphism}, \textit{Complexity}, \textit{Dialog Acts}, \textit{Emotions}, \textit{Politeness}, \textit{Rule-based Rewards} and \textit{Sycophancy} are all annotated using automated procedures. Overall, the non-automated feature blocks (\textit{User Margin}, \textit{Model Identity}, \textit{User}) are most important to our analysis. Of the automated feature blocks, most important appear to be those generated using external LLMs (\textit{Rule-based Rewards} and \textit{Sycophancy}) and \textit{Complexity}. The latter enjoys the benefit of years of validation in NLP/computational linguistics research, and the former are key contributions of prior published work with a minimal domain gap between their application and ours. Thus, ovoverall,e are reasonably confident that these features are accurate reflections of the constructs we aim to measure.

\section{Regression Analysis}
\label{app:regression}

As stated earlier, after regressing the set of annotations against the various endogenous variables, we are left with a $N \times D$ matrix of SHAP values.

SHAP values are the unique Shapley values of a conditional expectation function of the regression model. It provides, for each sample and feature, a scalar deviation from the grand regression model mean output:

\begin{equation}
    \hat{y}_{i}=\phi_{0}+\sum_{j=1}^{D}\phi_{ij}
\end{equation}

As such, we may interpret a SHAP value as the additive importance of a particular feature to the model output for a specific sample. The greater its magnitude, the greater its impact on the model's output in a particular direction. Given that we are analysing regression models trained on RM preference pairs, assuming the regression models are reasonable proxies for the RMs themselves, large positive SHAP values indicate large increases to the endogenous variable, whereas large negative SHAP values would indicate the opposite.

Thus, a SHAP value is a local explanation of model behavior (i.e., at preference pair-level). We are interested in global explanations (i.e., dataset-level). To convert from the former to the latter, we compute a number of summary statistics:
\begin{enumerate}
    \item \textit{MAV}: the arithmetic mean of the absolute value of the Shapley values, across all samples. The greater this value, the more extreme a features impact on average

    \item \textit{Correlation} ($r$): the Pearson correlation coefficient between the SHAP values and the exogenous feature. This captures the direction of the relationship between a feature's value and the expected SHAP value

    \item \textit{Beta} ($\beta$): defined as
        \begin{equation*}
            \beta=\mathtt{correlation}(\phi_{j}, \hat{y})\frac{\mathtt{std}(\hat{y})}{\mathtt{std}(\phi_{j})}
        \end{equation*}
    this captures the expected increase in Shapley values for a 1 unit increase in the exogenous variable. This is equivalent to the regression coefficient for a simple linear regression model. This metric captures both direction and magnitude; the greater this value, the more sensitive the model is to changes in the feature's value

    \item \textit{Harmonic Mean Rank}: the harmonic mean of the row-wise rank of the absolute SHAP values
        \begin{equation*}
            \left(\frac{1}{N}\sum_{i=1}^{N}\left(\mathtt{rank}(\phi_{i})_{j}\right
            )^{-1}\right)^{-1}
        \end{equation*}
    We only compute this metric whenever a feature is active (i.e., there is a difference between responses in a preference pair). Unlike the previous metrics, which capture the impact of a feature absolutely, this metric is relative to all other features active in a preference pair. 
    
\end{enumerate}

Thus, in order, we have summary statistics measuring the magnitude, the direction, the linear relationship, and the relative magnitude of a feature's importance. Of these, only the first (MAV) is commonly used; however, we found this inappropriate for evaluating features that are rarely active. 

We present SHAP values for individual features in Tables \ref{tab:regression_shap_individual_values} and \ref{tab:regression_community_individual_values}, which are located at the end of this paper for legibility reasons.

\begin{table*}[p]
    \centering
    \begin{tblr}{
    width = 1.0\textwidth,
    colspec = {lX[-1, c]X[-1, c]X[-1, c]X[-1, c]X[-1, c]X[-1, c]X[-1, c]X[-1, c]X[-1, c]},
    hline{1,3,Y,Z} = {solid},
    rowsep = 2.0pt,
    row{1} = {font=\bfseries},
    row{2} = {font=\itshape\small},
    cell{1}{1} = {r=2,c=1}{c},
    cell{1}{2} = {r=1,c=3}{c},
    cell{1}{5} = {r=1,c=3}{c},
    cell{1}{8} = {r=1,c=3}{c},
    row{3-Z} = {font=\small, abovesep = -2pt, belowsep = -2pt},
}
Block Name & \tm & & & \tg & & & \cm & & \\
& MAV ($\uparrow$) & HMR ($\downarrow$) & R2  ($\uparrow$) & MAV ($\uparrow$) & HMR ($\downarrow$) & R2  ($\uparrow$) & MAV ($\uparrow$) & HMR ($\downarrow$) & R2  ($\uparrow$) \\

Model & 0.47 & 1.41 & 0.27 & 0.44 & 1.50 & 0.12 & 0.45 & 1.28 & 0.02 \\
User Margin & 0.35 & 1.64 & 0.36 & 0.11 & 4.75 & 0.30 & 0.25 & 2.12 & 0.21 \\
Complexity & 0.15 & 3.19 & 0.00 & 0.36 & 1.83 & 0.00 & 0.20 & 2.58 & 0.01 \\
Rule-based Rewards & 0.12 & 3.77 & 0.00 & 0.25 & 2.52 & 0.00 & 0.13 & 3.63 & 0.00 \\
Emotions & 0.09 & 4.47 & 0.00 & 0.14 & 3.98 & 0.00 & 0.09 & 4.64 & 0.00 \\
Dialog Acts & 0.08 & 4.97 & 0.24 & 0.13 & 4.32 & 0.24 & 0.07 & 5.34 & 0.10 \\
User & 0.06 & 5.51 & 0.18 & 0.09 & 5.31 & 0.14 & 0.06 & 5.70 & 0.07 \\
Conversation & 0.05 & 6.01 & 0.22 & 0.06 & 6.64 & 0.22 & 0.06 & 5.79 & 0.09 \\
Sycophancy & 0.02 & 9.00 & 0.11 & 0.04 & 7.67 & 0.08 & 0.03 & 8.22 & 0.04 \\
Politeness & 0.03 & 7.44 & 0.00 & 0.05 & 7.10 & 0.00 & 0.02 & 8.06 & 0.00 \\
AnthroScore & 0.01 & 9.79 & 0.00 & 0.01 & 10.20 & 0.00 & 0.01 & 10.36 & 0.00 \\
PRISM & 0.00 & 11.32 & 0.00 & 0.01 & 10.79 & 0.00 & 0.00 & 10.77 & 0.00 \\

Total & & & 0.56 & & & 0.52 & & & 0.39 \\
\end{tblr}
    \caption{Summary metrics of block-wise net SHAP values for \prism.}
    \label{tab:regression_prism_shap_block}
\end{table*}

\begin{table*}[p]
    \centering
    \begin{tblr}{
    width = 1.0\textwidth,
    colspec = {lX[-1, c]X[-1, c]X[-1, c]X[-1, c]X[-1, c]X[-1, c]X[-1, c]X[-1, c]X[-1, c]},
    hline{1,3,Y,Z} = {solid},
    rowsep = 2.0pt,
    row{1} = {font=\bfseries},
    row{2} = {font=\itshape\small},
    cell{1}{1} = {r=2,c=1}{c},
    cell{1}{2} = {r=1,c=3}{c},
    cell{1}{5} = {r=1,c=3}{c},
    cell{1}{8} = {r=1,c=3}{c},
    row{3-Z} = {font=\small, abovesep = -2pt, belowsep = -2pt},
}
Block Name & \tm & & & \tg & & & \cm & & \\
& MAV ($\uparrow$) & HMR ($\downarrow$) & R2  ($\uparrow$) & MAV ($\uparrow$) & HMR ($\downarrow$) & R2  ($\uparrow$) & MAV ($\uparrow$) & HMR ($\downarrow$) & R2  ($\uparrow$) \\

User Margin & 0.83 & 1.10 & 0.47 & 0.38 & 2.11 & 0.13 & 0.47 & 1.20 & 0.24 \\
User & 0.28 & 1.89 & 0.35 & 0.50 & 1.36 & 0.03 & 0.19 & 2.09 & 0.19 \\
Emotions & 0.10 & 3.69 & 0.02 & 0.17 & 3.06 & 0.10 & 0.12 & 3.15 & 0.08 \\
Complexity & 0.10 & 3.63 & 0.00 & 0.20 & 2.69 & 0.01 & 0.11 & 3.32 & 0.00 \\
Dialog Acts & 0.07 & 4.42 & 0.15 & 0.13 & 3.85 & 0.14 & 0.08 & 4.14 & 0.15 \\
Politeness & 0.04 & 5.31 & 0.09 & 0.07 & 4.91 & 0.11 & 0.04 & 4.98 & 0.08 \\
Rule-based Rewards & 0.03 & 5.69 & 0.18 & 0.07 & 5.08 & 0.12 & 0.04 & 5.18 & 0.15 \\
Sycophancy & 0.01 & 7.08 & 0.02 & 0.02 & 7.36 & 0.03 & 0.01 & 7.50 & 0.02 \\
AnthroScore & 0.00 & 8.56 & 0.01 & 0.01 & 8.02 & 0.03 & 0.01 & 7.69 & 0.02 \\

Total & & & 0.63 & & & 0.39 & & & 0.34 \\
\end{tblr}
    \caption{Summary metrics of block-wise net SHAP values for \community.}
    \label{tab:regression_community_shap_block}
\end{table*}

\subsection{Block-Level Analysis}

For the purpose of reporting block-level importance, and to account for interference of different variables within the same block, we additionally compute block-level net SHAP values. We compute these by summing all SHAP values belonging to the same block, before computing the same summary statistics as before:
\begin{equation*}
    \phi^{\mathbb{F}}_{i}=\sum_{j\in \mathbb{F}}\phi_{ij}
\end{equation*}
where $\mathbb{F}$ is meant to denote a set of features belonging to the same block.

Beyond the discussed summary statistics, we also measure the ability of a block to predict the endogenous variable in isolation using the coefficient of determination ($R^{2}$).

We present these block-level SHAP values in Tables \ref{tab:regression_prism_shap_block} and \ref{tab:regression_community_shap_block}.

\subsection{Extended Feature-Level Analysis}

\begin{table*}[t]
    \begin{tblr}{
    width = \linewidth,
    colspec = {cX[-1, c]X[-1, c]X[-1, c]X[-1, c]X[-1, c]X[-1, c]},
    hline{1,2,Y,Z} = {solid},
    vline{Y} = {1-Z}{solid, abovepos = -1},
    row{1} = {font=\bfseries\small},
    row{2-Z} = {font=\small, abovesep = -2pt, belowsep = -2pt},
}
\diagbox[width=5em, height=3em]{Wave}{Educ}
& (At most) Complete Secondary & Some post-secondary & Post-secondary graduate & Some or complete graduate degree & Other & Total \\
1     & 0.02 & 0.06 & 0.42 & 0.37 & 0.00 & 0.87 \\
2     & 0.13 & 0.01 &      &      &      & 0.13 \\
Total & 0.14 & 0.07 & 0.42 & 0.37 & 0.00 & 1.00 \\
\end{tblr}

    \caption{A cross tabulation of the proportion of \community users belonging to each education level and collection wave.}
    \label{tab:data_community_educ_wave}
\end{table*}

\parheading{User} In Section \ref{sec:top_down}, we find that the RMs are memorizing a user metadata variable. Specifically, \cm is very high for user from the `(At most) Complete Secondary' education level. This is striking, since user metadata is never presented directly to the model. Instead, the model must infer this from the user's prompt and their response preference. Based on analyzing the prompts, we did not find a substantive difference in English proficiency, or the distribution of discussed topics. Instead, we find that the largest difference between these users and all other users in \community, is their collection wave. We show this in Table \ref{tab:data_community_educ_wave}. While \citeauthor{zhangCultivatingPluralismAlgorithmic2025} do not discuss the specific annotator recruitment process, they do briefly discuss the difficulty in constructing a representative sample of less educated users (see their Appendix D.3.1).

\parheading{Margin} \prism provides a scalar user-reported preference score for each model response, allowing us to express a margin of human preference for each preference pair. Unfortunately, \community only provides the index of the chosen response. They do, however, have instances where the same prompt is shown to different users, resulting in expression of contrasting personal preference. Thus, we compute two proxies for margin in user preference: 1) a preference pair-level margin in the probability of a response being selected for the same prompt (`Prob Margin'), and 2) a turn-level normalized entropy of the prompt-chosen response distribution (`Turn Entropy').

\parheading{Dataset Metadata} The \prism authors provide additional conversation- and response-level annotations. The latter correspond to cases where responses are non-English, contain Personally Identifiable Information (PII) or violate a safety policy. These cases are rare, resulting in very low block-level MAV scores (Figure \ref{fig:regression_prism_block_summary}), but yield large effects when active. Obvious safety policy violations make preference classification easier during both training and testing, whereas non-English responses see substantially decreased test-time performance without worse train-time performance, yielding high \cm. This suggests RMs have to memorize when users prefer non-English responses.

The provided annotations at the conversation-level---type and rating, `Conversation' in Figure \ref{fig:regression_prism_block_summary}---yield small to no difference in memorization or generalization scores. Controversy and values guided conversations tend to generate slightly lower \cm than unguided conversations. Individually, the conversation-level user ratings have minimal impact, although high `helpfulness' scores tend to decrease \cm. As such it is impossible to discern their importance, likely due to the high granularity of the annotations.

\begin{figure*}[p]
    \centering
    \includegraphics[width=1.0\textwidth]{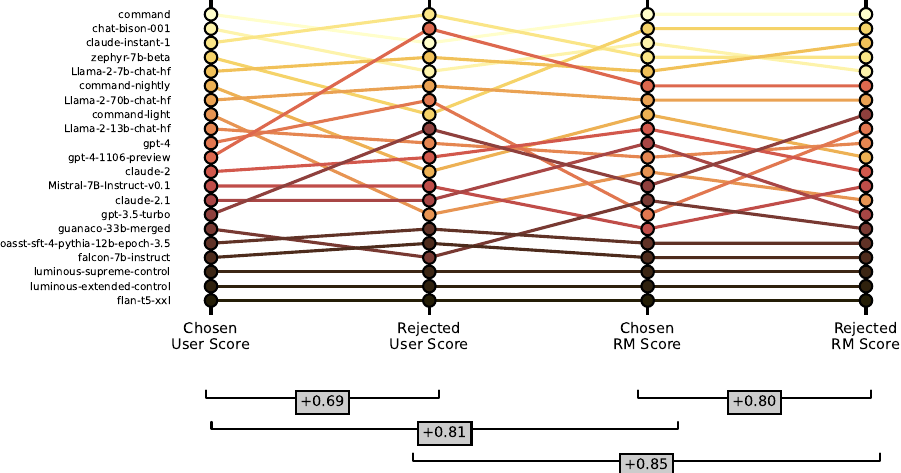}
    \caption{Correlation between the mean user score of a chosen and rejected response and the mean RM score of a chosen and rejected response. Model names are provided in text along the left-most vertical axis. Models are sorted by their mean `Chosen User Score', with the top being the most preferred and the bottom being the least preferred. Boxes at the bottom provide Kendall's Tau correlation value.}
    \label{fig:regression_prism_user_rm_preference_correlation}
\end{figure*}

\parheading{Model Identity} We present the mean user score and RM score for all \num{21} models used in \prism, conditioned on the model response being chosen or rejected, in Figure \ref{fig:regression_prism_user_rm_preference_correlation}. Users have a relatively consistent preference, with models whose responses receive high scores when chosen typically also receiving high scores when rejected ($\tau=$\qty{0.69}{\percent}). This preference is even more consistent in the RM preference ($\tau=$\qty{0.80}{\percent}), especially most and least preferred models. Comparing users against RMs, we see that their preference is strongly associated with each other ($\tau=$\qtyrange{0.81}{0.85}{\percent}).

\begin{figure*}[t]
    \centering
    \begin{subfigure}{1.0\textwidth}
        \includegraphics[width=\textwidth]{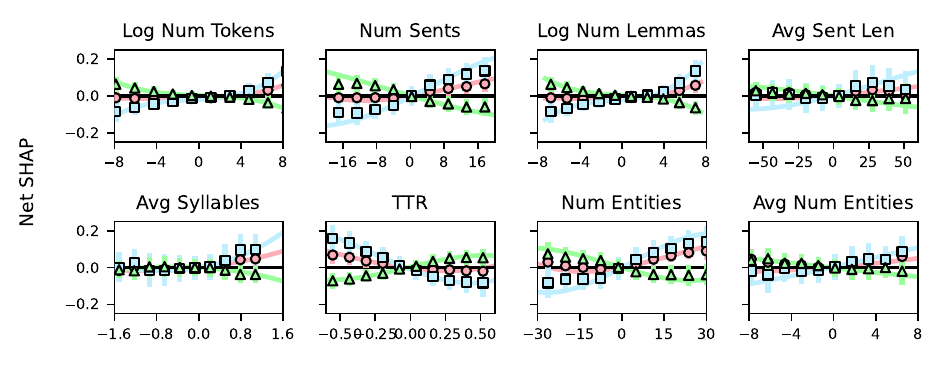}
        \caption{\prism}
    \end{subfigure}
    \\
    \begin{subfigure}{1.0\textwidth}
        \includegraphics[width=\textwidth]{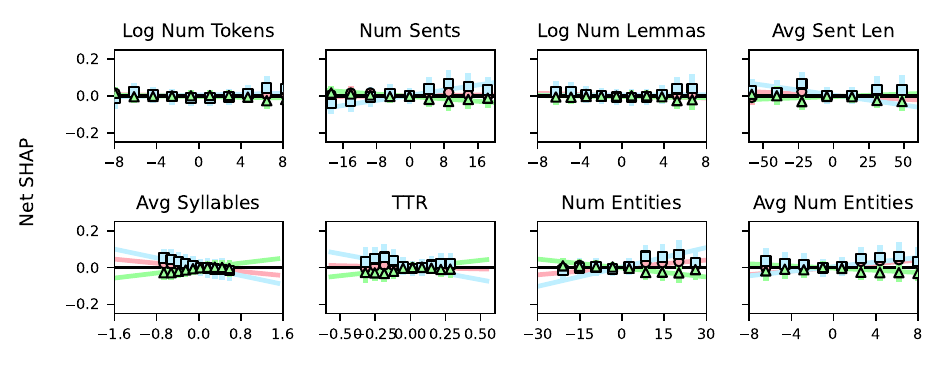}
        \caption{\community}
    \end{subfigure}
    \caption{\textbf{Complexity partial effect}: block-level net SHAP values induced by complexity differences between responses. Shapes indicate mean per bin, whereas solid lines give linear trend lines. We trim the data bins between the 0.1st and 99.9th percentiles.}
    \label{fig:regression_complexity_partial_effect}
    \begin{subfigure}{0.48\textwidth}
        \includegraphics[width=\textwidth]{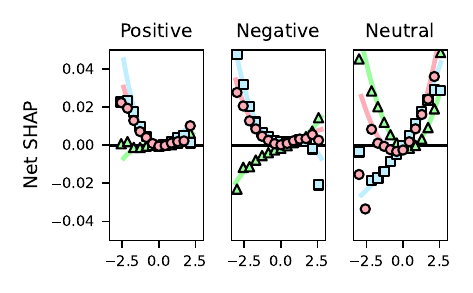}
        \caption{\prism}
    \end{subfigure}
    \begin{subfigure}{0.48\textwidth}
        \includegraphics[width=\textwidth]{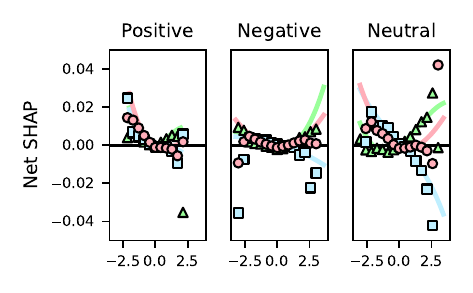}
        \caption{\community}
    \end{subfigure}
    \caption{\textbf{Emotion partial effect}: block-level net SHAP values induced by emotion differences between responses. We aggregate SHAP scores based on sentiment categories, as described in \citet{demszkyGoEmotionsDatasetFineGrained2020}'s Figure 2. Solid lines now give the quadratic trend line.}
    \label{fig:regression_emotion_partial_effect}
\end{figure*}

\parheading{Complexity} We graphically depict the effect of increasing length, complexity and diversity on the expected block-level net SHAP values in Figure \ref{fig:regression_complexity_partial_effect}. In general, these have a stronger positive correlation with \tg than with \tm, resulting in negative \cm when the chosen response is much longer or more complex than the rejected response, and positive \cm whenever the opposite occurs. The effect is fairly linear, being much larger at the extremes than near null difference. The major exception to this finding is \texttt{TTR}, a measure of token diversity in the responses, with \textit{less} diverse chosen responses resulting in better performance. We posit that this is due to noise outliers, where the rejected response contains long but random tokens, resulting in diverse but meaningless text.

\parheading{Emotion} We similarly graphically display the effect of emotion (aggregated into sentiment bins) on the expected block-level net SHAP values in Figure \ref{fig:regression_emotion_partial_effect}. Unlike complexity, the trend does not seem to be linear, with human preference classification tending to be easier when extreme emotional differences are present. Overall, however, emotion tends to be modestly predictive of either generalization or memorization in RMs.

\section{SAE Analysis}
\label{app:sae}

To generate a set of data-driven \cm determinants, we reimplement the SAEs used by \citet{movvaWhatsMyHuman2025}. They train these on the difference of response-level embeddings, requiring the SAE to reconstruct the difference embedding. They use OpenAI \texttt{text-embedding-3-small} as the embedding model, whereas we use \href{https://huggingface.co/meta-llama/Llama-3.2-1B}{\texttt{Llama-3.2-1B}}, the same checkpoint from which we initialize the RMs, ensuring the features are generated from information which is---in theory---available to the RMs as well. We additionally experiment with several SoTA RMs and general text embedding models, but these did not produce more interpretable features.

For the SAE instantiation, we use the exact same Matryoshka Batch Top-k \cite{bussmannBatchTopKSparseAutoencoders2024, bussmannLearningMultiLevelFeatures2025}, with a latent space of \num{32} and at most \num{4} active features per sample. To automatically produce an interpretation of the SAE's latent space, we sample \num{5} preference pairs from the \num{95}th percentile or above, and have OpenAI \texttt{gpt-5} produce a succinct summary of the most salient difference between the responses. We repeat this \num{5} times for each description. Finally, to validate the description, we additionally annotate \num{300} response pairs for said feature, sampled uniformly across the feature's latent space embedding, and have OpenAI \texttt{gpt-5-nano} annotate these for said feature. We then only keep the description that maximally correlates with the latent space values.

Regardless of the specific response embedding model or SAE hyperparameters, we find it difficult to replicate the features reported by \citeauthor{movvaWhatsMyHuman2025} (see their Appendix E Tables 10 \& 12). We find substantial multi-collinearity between the latent dimensions of the SAE, which translates to near identical feature interpretations. Furthermore, we find that the problem is exacerbated when the SAE is trained on each \cm region in isolation.

To generate features that are simultaneously interpretable and discriminative, we add an auxilliary \cm region classification head on the sparsified SAE latent space. We find that equally weighting the reconstruction and classification losses works optimally, and minimally reduces the ability of the SAE to reconstruct the difference embeddings. The regions are displayed in Figure \ref{fig:hook}, and Figure \ref{fig:sae_discriminative_regions}.

We present all valid feature descriptions for both datasets in Table \ref{tab:sae_features}. We additionally provide the positive ($\frac{1}{N}\sum_{i=1}^{N} 1(z_{i,j}>c_{j})$) and negative ($\frac{1}{N}\sum_{i=1}^{N} 1(z_{i,j}<-c_{j})$) prevalence of the feature (where $c_{j}$ is the learned SAE sparsification threshold; Pearson's correlation coefficient between validation samples and the latent feature values, and its associated $p$-value, and; the classifier weights between the SAE feature and each \cm region. The greater this weight is, the greater the positive association between this feature and a corresponding region.

\section{Individual SHAP Value Tables}
\label{app:regression_individual}

We present all SHAP values for individual features in Tables \ref{tab:regression_shap_individual_values} and \ref{tab:regression_community_individual_values}. We omit the model identity values for brevity, but these are displayed graphically in Figure \ref{fig:regression_prism_model_identity}.

\clearpage

\vfill

\begin{table*}
    \centering
\begin{tblr}{
    width = 1.0\textwidth,
    colspec = {X[1, c]X[1, c]X[1, c]X[2, c]X[1, c]X[1, c]X[1, c]X[1, c]X[11, l]},
    hline{1,3,Z} = {solid},
    row{1} = {font=\bfseries},
    row{2} = {font=\itshape\small},
    cell{1}{1} = {r=1,c=2}{c},
    cell{1}{3} = {r=1,c=2}{c},
    cell{1}{5} = {r=1,c=4}{c},
    cell{1}{9} = {r=2,c=1}{c},
    cell{3}{1} = {r=1,c=9}{c},
    cell{26}{1} = {r=1,c=9}{c},
    row{3-Z} = {font=\small, abovesep = -2pt, belowsep = -2pt},
    colsep=1pt,
}

Prev. & & Corr. & & Weight & & & & Description \\
+ & - & $r$ & $p$ & 0 & 1 & 2 & 3 & \\

\prism \\
\SetHline{1-Z}{lightgray, 0.5pt, solid}
0\% & 1\% & 0.79 & 6.36e-66 & 0.03 & -0.07 & 0.17 & 0.09 & gives only a one-word affirmation \\
2\% & 0\% & 0.72 & 9.90e-50 & 0.20 & -0.31 & 0.13 & -0.16 & provides real-time weather forecasts/details \\
0\% & 1\% & 0.72 & 5.97e-49 & 0.02 & -0.02 & -0.10 & 0.03 & responds in a non-English language \\
1\% & 6\% & 0.71 & 1.21e-46 & -0.08 & -0.00 & -0.06 & -0.10 & gives terse one-word reply without elaboration \\
1\% & 1\% & 0.62 & 4.69e-33 & 0.05 & -0.10 & 0.13 & 0.03 & provides detailed recipe ingredients and steps \\
0\% & 0\% & 0.59 & 5.00e-29 & 0.03 & 0.07 & 0.12 & -0.05 & is an empty (blank) response \\
0\% & 0\% & 0.57 & 3.43e-27 & 0.05 & -0.05 & 0.08 & 0.23 & is written in English \\
14\% & 18\% & 0.57 & 4.74e-27 & -0.06 & 0.02 & -0.19 & 0.15 & does not self-identify as an AI model \\
1\% & 1\% & 0.52 & 2.28e-22 & 0.01 & -0.23 & -0.09 & 0.06 & only a generic greeting \\
1\% & 2\% & 0.48 & 1.71e-18 & -0.08 & -0.17 & -0.06 & 0.22 & avoids specific, detailed meal recommendations \\
0\% & 100\% & 0.41 & 8.34e-14 & 0.18 & -0.07 & 0.50 & -0.46 & gives brief, non-elaborated responses \\
4\% & 5\% & 0.41 & 1.09e-13 & -0.07 & 0.07 & -0.18 & 0.02 & includes explicit AI self-identification disclaimers \\
15\% & 10\% & 0.35 & 2.65e-10 & 0.01 & -0.16 & 0.10 & -0.18 & provides direct substantive explanation or definition \\
6\% & 5\% & 0.31 & 3.78e-08 & -0.05 & -0.05 & -0.18 & 0.11 & gives detailed, actionable step-by-step guidance \\
2\% & 6\% & 0.30 & 7.10e-08 & -0.19 & 0.42 & 0.10 & -0.16 & states a definite stance on sensitive issues \\
1\% & 1\% & 0.28 & 1.01e-06 & 0.24 & -0.24 & 0.05 & 0.24 & avoids giving political opinions or analysis \\
9\% & 16\% & 0.27 & 1.90e-06 & 0.00 & -0.05 & -0.04 & 0.15 & terse or empty reply without elaboration \\
10\% & 4\% & 0.26 & 3.41e-06 & 0.03 & -0.47 & -0.09 & 0.19 & provides an on-topic answer instead of refusing \\
2\% & 10\% & 0.26 & 4.76e-06 & -0.21 & 0.68 & 0.06 & -0.24 & does not claim personal preferences or experiences \\
6\% & 1\% & 0.23 & 4.18e-05 & 0.36 & -1.00 & -0.13 & 0.25 & does not directly answer the prompt \\
13\% & 16\% & 0.23 & 4.69e-05 & -0.08 & 0.08 & -0.16 & 0.19 & does not use AI-disclaimer neutrality hedging \\
2\% & 5\% & 0.22 & 1.32e-04 & -0.20 & 0.64 & -0.07 & -0.41 & gives detailed factual background on conflict \\
\SetHline{1-Z}{lightgray, 0.5pt, solid}
\community \\
\SetHline{1-Z}{lightgray, 0.5pt, solid}
33\% & 7\% & 0.75 & 1.14e-54 & 0.38 & -0.21 & -0.01 & -0.18 & narrative paragraph format without bullet lists \\
35\% & 6\% & 0.71 & 8.18e-48 & 0.52 & -0.36 & -0.10 & -0.21 & written as narrative paragraphs, not a list \\
49\% & 2\% & 0.69 & 3.14e-43 & 0.40 & -0.36 & -0.35 & 0.23 & uses structured bullet-point lists/templates \\
2\% & 10\% & 0.65 & 1.67e-37 & -0.11 & 0.14 & 0.26 & -0.07 & written as continuous prose, not a list \\
53\% & 2\% & 0.64 & 4.07e-36 & 0.40 & -0.46 & -0.30 & -0.20 & uses structured outline/list format \\
1\% & 12\% & 0.61 & 9.40e-32 & -0.67 & 0.65 & 0.60 & -0.26 & uses structured, formatted documents or lists \\
1\% & 3\% & 0.57 & 1.28e-27 & 0.13 & 0.08 & 0.09 & -0.08 & emphasizes sustainability and environmental impact \\
2\% & 4\% & 0.53 & 5.47e-23 & -0.07 & 0.20 & 0.15 & -0.33 & abstract, long-form philosophical exposition \\
5\% & 5\% & 0.49 & 1.89e-19 & 0.05 & -0.33 & -0.13 & 0.23 & focuses on evocative, experiential descriptions, not logistics \\
1\% & 2\% & 0.45 & 1.05e-16 & 0.14 & -0.10 & -0.19 & 0.22 & emphasizes technological/AI-driven solutions \\
3\% & 2\% & 0.43 & 1.16e-14 & 0.02 & -0.00 & -0.26 & 0.20 & does not emphasize DEI or social justice \\
0\% & 0\% & 0.40 & 7.52e-13 & 0.28 & 0.11 & -0.02 & -0.27 & prose narrative, not screenplay format \\
2\% & 1\% & 0.39 & 3.09e-12 & 0.09 & 0.05 & -0.21 & 0.03 & does not emphasize historical/heritage content \\
100\% & 0\% & 0.38 & 1.20e-11 & -0.22 & -0.31 & -0.69 & 0.47 & uses a formal, elevated tone \\
0\% & 1\% & 0.38 & 1.45e-11 & -0.29 & 0.39 & 0.15 & -0.32 & does not include a sample template \\
3\% & 3\% & 0.37 & 6.62e-11 & 0.35 & -0.12 & -0.18 & -0.06 & uses descriptive long-form prose instead of lists \\
1\% & 1\% & 0.36 & 1.31e-10 & -0.03 & -0.06 & 0.16 & 0.09 & emphasizes direct interpersonal communication and support \\
1\% & 1\% & 0.36 & 1.97e-10 & 0.11 & 0.21 & 0.26 & -0.14 & does not emphasize mindfulness or holistic self-care \\
1\% & 0\% & 0.35 & 3.55e-10 & 0.04 & -0.45 & -0.33 & 0.25 & focuses on policies, safeguards, and official processes \\
1\% & 2\% & 0.33 & 5.67e-09 & -0.06 & 0.00 & -0.09 & 0.15 & avoids time-management and scheduling advice \\
1\% & 1\% & 0.28 & 6.91e-07 & -0.13 & 0.06 & 0.10 & -0.35 & interprets subjects in spiritual/mythic terms \\
0\% & 1\% & 0.28 & 7.20e-07 & -0.05 & -0.18 & 0.04 & 0.12 & avoids first-person capability statements \\
1\% & 1\% & 0.27 & 1.50e-06 & 0.15 & -0.44 & -0.39 & 0.33 & does not emphasize luxury or exclusivity \\
2\% & 1\% & 0.27 & 1.81e-06 & 0.30 & -0.23 & -0.22 & 0.20 & emphasizes urban comforts and cultural amenities \\
1\% & 0\% & 0.26 & 4.63e-06 & 0.24 & -0.07 & -0.05 & 0.03 & focuses on functional, technical specifications \\
3\% & 7\% & 0.24 & 2.64e-05 & -0.18 & 0.06 & 0.15 & -0.07 & gives high-level guidance, not itemized lists \\
1\% & 1\% & 0.24 & 3.67e-05 & -0.04 & 0.39 & 0.41 & -0.23 & prioritizes balance, adaptability, and self-care \\
3\% & 1\% & 0.22 & 1.24e-04 & 0.25 & -0.37 & -0.38 & 0.23 & recommends conventional, non-immersive options \\
1\% & 2\% & 0.20 & 5.99e-04 & -0.17 & 0.26 & 0.26 & -0.27 & uses a formal, impersonal tone \\

\end{tblr}
    \caption{Discovered conditional SAE features. Prevalence gives the proportion of preference pairs for which the feature is positively (+) or negatively (-) active. Correlation gives the Pearson correlation coefficient and corresponding $p$ value between the SAE's latent dimension and LLM annotations for the description. The weights columns give the conditional association weights between the latent feature and each CM region.}
    \label{tab:sae_features}
\end{table*}

\clearpage

\onecolumn
\begin{longtblr}[
    caption = {Summary statistics for individual SHAP values on \prism. We exclude model identity fixed effects.},
    label = {tab:regression_shap_individual_values},
]{
    width = 1.0\textwidth,
    colspec = {X[c]X[l]cccccccccccc},
    colsep = 2.0pt,
    rowhead = 2,
    hline{1, 3, Z} = {solid},
    cells = {
        font=\footnotesize
    },
    row{1} = {font=\bfseries},
    row{2} = {font=\itshape\small},
    cell{1}{1} = {r=2,c=1}{c},
    cell{1}{2} = {r=2,c=1}{c},
    cell{1}{3} = {r=2,c=1}{c},
    cell{1}{4} = {r=1,c=4}{c},
    cell{1}{8} = {r=1,c=4}{c},
    cell{1}{12} = {r=1,c=4}{c},
    row{3-Z} = {
        font = \tiny,
        rowsep = 0pt,
        abovesep = -3pt,
        belowsep = -3pt,
    },
    cell{3}{1} = {r=4,c=1}{c},
    cell{7}{1} = {r=3,c=1}{c},
    cell{10}{1} = {r=7,c=1}{c},
    cell{17}{1} = {r=6,c=1}{c},
    cell{23}{1} = {r=9,c=1}{c},
    cell{32}{1} = {r=8,c=1}{c},
    cell{40}{1} = {r=5,c=1}{c},
    cell{45}{1} = {r=9,c=1}{c},
    cell{54}{1} = {r=4,c=1}{c},
    cell{58}{1} = {r=3,c=1}{c},
    cell{61}{1} = {r=5,c=1}{c},
    cell{66}{1} = {r=11,c=1}{c},
    cell{77}{1} = {r=5,c=1}{c},
    cell{82}{1} = {r=12,c=1}{c},
    cell{94}{1} = {r=1,c=1}{c},
    cell{95}{1} = {r=8,c=1}{c},
    cell{103}{1} = {r=23,c=1}{c},
    cell{126}{1} = {r=28,c=1}{c},
    cell{154}{1} = {r=4,c=1}{c},
    cell{158}{1} = {r=15,c=1}{c},
    cell{173}{1} = {r=3,c=1}{c},
}

Block & Name & Prop & \tm & & & & \tg & & & & \cm \\
                & & & MAV & r & $\beta$ & HMR & MAV & r & $\beta$ & HMR & MAV & r & $\beta$ & HMR \\

Turn & User Margin & 100.00\% & 0.35 & +0.98 & +0.45 & 1.51 & 0.11 & +0.96 & +0.14 & 5.83 & 0.25 & +0.96 & +0.30 & 1.90 \\
Turn & English Flag & 1.23\% & 0.00 & +0.24 & +0.01 & 43.78 & 0.00 & -0.94 & -0.13 & 9.25 & 0.00 & +0.87 & +0.10 & 8.10 \\
Turn & PII Flag & 2.34\% & 0.00 & -0.34 & -0.01 & 25.49 & 0.00 & +0.74 & +0.04 & 17.47 & 0.00 & -0.72 & -0.04 & 12.14 \\
Turn & Moderation Flag & 0.57\% & 0.00 & +0.99 & +0.21 & 2.83 & 0.00 & +0.97 & +0.23 & 4.08 & 0.00 & +0.81 & +0.06 & 12.52 \\

\SetHline{1-Z}{lightgray, 0.5pt, solid}
Conversation & Unguided & 39.27\% & 0.01 & +0.47 & +0.01 & 48.40 & 0.02 & -0.69 & -0.04 & 33.52 & 0.02 & +0.91 & +0.04 & 22.49 \\
Conversation / Type & Controversy Guided & 30.11\% & 0.03 & -0.90 & -0.07 & 12.03 & 0.02 & -0.33 & -0.02 & 37.37 & 0.02 & -0.79 & -0.05 & 17.31 \\
Conversation & Values Guided & 30.62\% & 0.02 & -0.95 & -0.05 & 17.06 & 0.01 & -0.51 & -0.02 & 46.89 & 0.01 & -0.72 & -0.02 & 40.64 \\

\SetHline{1-Z}{lightgray, 0.5pt, dashed}
Conversation / Rating & Helpfulness & 99.39\% & 0.01 & +0.08 & +0.00 & 45.55 & 0.02 & +0.74 & +0.02 & 46.14 & 0.02 & -0.92 & -0.03 & 27.29 \\
Conversation / Rating & Creativity & 89.81\% & 0.02 & +0.87 & +0.03 & 22.28 & 0.02 & +0.31 & +0.01 & 47.89 & 0.01 & +0.68 & +0.02 & 39.11 \\
Conversation / Rating & Diversity & 88.45\% & 0.01 & -0.72 & -0.01 & 44.15 & 0.02 & -0.48 & -0.01 & 49.88 & 0.01 & +0.02 & +0.00 & 43.89 \\
Conversation / Rating & Safety & 94.29\% & 0.01 & -0.54 & -0.01 & 54.47 & 0.02 & +0.24 & +0.01 & 51.42 & 0.01 & -0.73 & -0.01 & 47.99 \\
Conversation / Rating & Values & 88.64\% & 0.01 & -0.71 & -0.01 & 39.94 & 0.02 & -0.43 & -0.02 & 37.53 & 0.01 & -0.40 & -0.01 & 54.69 \\
Conversation / Rating & Factuality & 97.96\% & 0.01 & +0.58 & +0.01 & 47.25 & 0.02 & +0.40 & +0.01 & 49.56 & 0.01 & -0.21 & -0.00 & 62.32 \\
Conversation / Rating & Fluency & 99.62\% & 0.02 & -0.91 & -0.03 & 28.11 & 0.02 & -0.64 & -0.02 & 51.01 & 0.01 & -0.48 & -0.01 & 67.20 \\

\SetHline{1-Z}{lightgray, 0.5pt, solid}
User / Age & 18-24 & 20.11\% & 0.01 & -0.83 & -0.03 & 31.35 & 0.01 & -0.55 & -0.02 & 39.34 & 0.01 & -0.69 & -0.01 & 55.91 \\
User / Age & 25-34 & 31.01\% & 0.01 & -0.79 & -0.03 & 25.03 & 0.02 & -0.72 & -0.04 & 29.70 & 0.00 & -0.17 & -0.00 & 66.22 \\
User / Age & 35-44 & 16.08\% & 0.00 & -0.58 & -0.01 & 57.39 & 0.01 & -0.57 & -0.02 & 43.01 & 0.00 & +0.65 & +0.01 & 47.88 \\
User / Age & 45-54 & 13.54\% & 0.00 & +0.03 & +0.00 & 39.73 & 0.01 & -0.20 & -0.01 & 41.16 & 0.00 & +0.14 & +0.00 & 47.19 \\
User / Age & 55-65 & 12.87\% & 0.00 & +0.45 & +0.01 & 39.26 & 0.01 & +0.00 & +0.00 & 46.07 & 0.00 & +0.62 & +0.02 & 27.16 \\
User / Age & 65+ & 6.39\% & 0.00 & -0.79 & -0.01 & 39.48 & 0.01 & -0.72 & -0.04 & 21.29 & 0.00 & +0.89 & +0.03 & 18.59 \\

\SetHline{1-Z}{lightgray, 0.5pt, dashed}
User / Education & Some Primary & 0.24\% & 0.00 & -0.99 & -0.15 & 3.29 & 0.00 & -0.93 & -0.12 & 5.10 & 0.00 & +0.45 & +0.01 & 41.42 \\
User / Education & Completed Primary School & 0.96\% & 0.00 & +0.72 & +0.03 & 13.19 & 0.00 & +0.89 & +0.10 & 6.22 & 0.00 & -0.92 & -0.08 & 6.85 \\
User / Education & Some Secondary & 1.54\% & 0.00 & +0.96 & +0.12 & 4.45 & 0.00 & +0.71 & +0.05 & 15.54 & 0.00 & +0.92 & +0.09 & 7.05 \\
User / Education & Secondary & 13.78\% & 0.01 & -0.92 & -0.06 & 11.10 & 0.01 & -0.74 & -0.04 & 25.91 & 0.00 & +0.06 & +0.00 & 43.09 \\
User / Education & Some University & 15.49\% & 0.01 & +0.84 & +0.02 & 31.13 & 0.01 & +0.63 & +0.03 & 27.17 & 0.01 & +0.18 & +0.00 & 37.01 \\
User / Education & Vocational & 8.10\% & 0.00 & -0.65 & -0.02 & 32.71 & 0.01 & -0.53 & -0.02 & 29.73 & 0.00 & +0.48 & +0.02 & 29.26 \\
User / Education & Bachelors Degree & 42.66\% & 0.01 & -0.50 & -0.01 & 65.17 & 0.02 & -0.44 & -0.02 & 62.90 & 0.01 & +0.55 & +0.01 & 57.36 \\
User / Education & Professional Degree & 16.78\% & 0.01 & +0.75 & +0.02 & 25.70 & 0.01 & -0.55 & -0.02 & 47.29 & 0.01 & +0.94 & +0.05 & 14.03 \\
User / Education & Prefer not to Say & 0.45\% & 0.00 & -0.80 & -0.04 & 14.76 & 0.00 & -0.79 & -0.07 & 12.86 & 0.00 & +0.22 & +0.01 & 18.00 \\

\SetHline{1-Z}{lightgray, 0.5pt, dashed}
User / Employment & Student & 12.67\% & 0.01 & -0.67 & -0.02 & 31.51 & 0.01 & -0.88 & -0.06 & 17.09 & 0.01 & +0.93 & +0.04 & 16.41 \\
User / Employment & Unemployed/Not looking & 3.27\% & 0.01 & +0.95 & +0.07 & 7.34 & 0.00 & +0.47 & +0.02 & 18.53 & 0.01 & +0.91 & +0.07 & 8.64 \\
User / Employment & Unemployed/Looking & 7.94\% & 0.01 & -0.87 & -0.05 & 12.47 & 0.01 & -0.60 & -0.03 & 22.98 & 0.00 & -0.27 & -0.01 & 31.65 \\
User / Employment & Homemaker & 3.07\% & 0.00 & -0.96 & -0.07 & 7.81 & 0.00 & -0.86 & -0.07 & 8.53 & 0.00 & +0.18 & +0.00 & 43.10 \\
User / Employment & Working/Part-time & 17.48\% & 0.00 & +0.15 & +0.00 & 43.59 & 0.01 & +0.29 & +0.01 & 49.43 & 0.00 & -0.21 & -0.00 & 54.63 \\
User / Employment & Working/Full-time & 47.53\% & 0.01 & -0.62 & -0.01 & 79.99 & 0.01 & -0.34 & -0.01 & 71.62 & 0.01 & +0.44 & +0.01 & 61.86 \\
User / Employment & Retired & 6.68\% & 0.00 & +0.33 & +0.01 & 35.25 & 0.00 & -0.11 & -0.00 & 45.90 & 0.00 & +0.35 & +0.01 & 33.68 \\
User / Employment & Prefer not to Say & 1.36\% & 0.00 & +0.62 & +0.01 & 29.60 & 0.00 & +0.93 & +0.11 & 5.34 & 0.00 & -0.89 & -0.06 & 8.86 \\

\SetHline{1-Z}{lightgray, 0.5pt, dashed}
User / English Proficiency & Basic & 0.30\% & 0.00 & -0.97 & -0.16 & 3.53 & 0.00 & -0.94 & -0.20 & 2.61 & 0.00 & +0.90 & +0.10 & 6.48 \\
User / English Proficiency & Intermediate & 2.80\% & 0.00 & -0.50 & -0.01 & 41.63 & 0.00 & -0.44 & -0.02 & 26.53 & 0.00 & +0.42 & +0.02 & 22.19 \\
User / English Proficiency & Advanced & 10.85\% & 0.00 & -0.85 & -0.02 & 29.17 & 0.01 & -0.84 & -0.06 & 15.83 & 0.00 & +0.51 & +0.01 & 34.26 \\
User / English Proficiency & Fluent & 26.84\% & 0.01 & -0.86 & -0.04 & 17.02 & 0.01 & -0.33 & -0.01 & 44.18 & 0.01 & -0.78 & -0.04 & 19.45 \\
User / English Proficiency & Native & 59.22\% & 0.02 & -0.85 & -0.04 & 63.69 & 0.01 & -0.55 & -0.02 & 99.90 & 0.01 & -0.81 & -0.02 & 99.72 \\

\SetHline{1-Z}{lightgray, 0.5pt, dashed}
User / Ethnicity & Asian & 6.46\% & 0.00 & +0.60 & +0.02 & 21.77 & 0.00 & -0.33 & -0.02 & 26.99 & 0.01 & +0.91 & +0.05 & 12.60 \\
User / Ethnicity & Black & 7.59\% & 0.00 & -0.81 & -0.01 & 46.34 & 0.00 & -0.53 & -0.02 & 40.12 & 0.00 & +0.67 & +0.02 & 41.33 \\
User / Ethnicity & Hispanic & 8.50\% & 0.00 & -0.43 & -0.01 & 40.88 & 0.00 & +0.47 & +0.02 & 39.96 & 0.00 & +0.34 & +0.01 & 28.20 \\
User / Ethnicity & Indigenous & 0.54\% & 0.00 & -0.03 & -0.00 & 13.10 & 0.00 & -0.38 & -0.03 & 18.75 & 0.00 & +0.24 & +0.01 & 19.55 \\
User / Ethnicity & Middle Eastern & 0.92\% & 0.00 & -0.92 & -0.06 & 9.05 & 0.00 & +0.56 & +0.04 & 20.11 & 0.00 & -0.93 & -0.08 & 7.38 \\
User / Ethnicity & Mixed & 4.94\% & 0.00 & -0.61 & -0.02 & 25.99 & 0.00 & -0.68 & -0.03 & 27.76 & 0.00 & +0.67 & +0.03 & 19.55 \\
User / Ethnicity & White & 65.19\% & 0.01 & -0.66 & -0.02 & 77.71 & 0.02 & -0.49 & -0.02 & 75.70 & 0.01 & +0.21 & +0.00 & 92.11 \\
User / Ethnicity & Other & 1.03\% & 0.00 & -0.58 & -0.02 & 19.28 & 0.00 & +0.13 & +0.01 & 22.43 & 0.00 & -0.62 & -0.03 & 17.17 \\
User / Ethnicity & Prefer not to Say & 4.82\% & 0.00 & -0.50 & -0.01 & 45.58 & 0.00 & -0.56 & -0.03 & 25.30 & 0.00 & +0.63 & +0.02 & 25.54 \\

\SetHline{1-Z}{lightgray, 0.5pt, dashed}
User / Gender & Female & 47.94\% & 0.01 & -0.89 & -0.03 & 50.64 & 0.01 & -0.62 & -0.02 & 75.98 & 0.01 & -0.81 & -0.02 & 65.37 \\
User / Gender & Male & 50.42\% & 0.04 & -0.92 & -0.07 & 14.56 & 0.03 & -0.82 & -0.06 & 32.07 & 0.01 & -0.55 & -0.01 & 66.73 \\
User / Gender & Non-binary / Third & 1.58\% & 0.00 & -0.74 & -0.02 & 23.83 & 0.00 & -0.53 & -0.02 & 29.71 & 0.00 & +0.75 & +0.03 & 20.94 \\
User / Gender & Prefer not to Say & 0.07\% & 0.00 & -0.56 & -0.03 & 12.95 & 0.00 & -0.22 & -0.02 & 14.27 & 0.00 & -0.73 & -0.05 & 9.58 \\

\pagebreak

\SetHline{1-Z}{lightgray, 0.5pt, dashed}
User / LLM Familiarity & Not familiar & 9.78\% & 0.01 & -0.85 & -0.05 & 13.73 & 0.01 & -0.73 & -0.04 & 19.87 & 0.00 & -0.10 & -0.00 & 56.48 \\
User / LLM Familiarity & Somewhat & 62.13\% & 0.02 & -0.91 & -0.04 & 41.91 & 0.02 & -0.72 & -0.04 & 51.26 & 0.01 & -0.58 & -0.01 & 89.38 \\
User / LLM Familiarity & Very & 28.08\% & 0.02 & -0.95 & -0.06 & 13.54 & 0.02 & -0.87 & -0.05 & 28.67 & 0.01 & -0.70 & -0.02 & 39.22 \\

\SetHline{1-Z}{lightgray, 0.5pt, dashed}
User / LLM Frequency & < Once a Year & 10.50\% & 0.00 & -0.01 & -0.00 & 56.09 & 0.00 & -0.01 & -0.00 & 59.85 & 0.00 & -0.14 & -0.00 & 39.16 \\
User / LLM Frequency & Once per Month & 26.30\% & 0.01 & -0.72 & -0.01 & 58.55 & 0.01 & -0.70 & -0.03 & 40.55 & 0.01 & +0.40 & +0.01 & 47.41 \\
User / LLM Frequency & > Once a Month & 19.27\% & 0.01 & -0.70 & -0.01 & 51.26 & 0.01 & -0.25 & -0.01 & 55.06 & 0.01 & -0.53 & -0.02 & 42.13 \\
User / LLM Frequency & Every Week & 20.45\% & 0.01 & -0.48 & -0.01 & 28.49 & 0.01 & +0.77 & +0.03 & 26.40 & 0.02 & -0.92 & -0.05 & 13.75 \\
User / LLM Frequency & Every Day & 7.30\% & 0.01 & -0.91 & -0.04 & 13.26 & 0.01 & -0.71 & -0.04 & 22.00 & 0.00 & -0.60 & -0.01 & 49.60 \\

\SetHline{1-Z}{lightgray, 0.5pt, dashed}
User / Location & Sub-Saharan Africa & 5.28\% & 0.00 & +0.48 & +0.02 & 14.13 & 0.00 & -0.72 & -0.04 & 25.52 & 0.01 & +0.95 & +0.08 & 6.95 \\
User / Location & Northern America & 27.91\% & 0.01 & -0.40 & -0.01 & 56.94 & 0.01 & -0.71 & -0.03 & 38.30 & 0.01 & +0.44 & +0.01 & 47.17 \\
User / Location & Latin America & 9.28\% & 0.00 & -0.59 & -0.02 & 32.06 & 0.00 & -0.24 & -0.01 & 42.43 & 0.00 & -0.12 & -0.00 & 47.86 \\
User / Location & Eastern Asia & 1.25\% & 0.00 & -0.41 & -0.01 & 27.53 & 0.00 & -0.69 & -0.07 & 10.34 & 0.00 & +0.80 & +0.04 & 14.70 \\
User / Location & Western Asia & 3.85\% & 0.00 & -0.71 & -0.02 & 25.67 & 0.01 & -0.96 & -0.13 & 5.63 & 0.01 & +0.95 & +0.11 & 4.81 \\
User / Location & Aus \& NZ & 10.17\% & 0.00 & +0.60 & +0.01 & 32.55 & 0.01 & +0.42 & +0.02 & 36.48 & 0.00 & +0.63 & +0.02 & 25.20 \\
User / Location & Northern Europe & 29.70\% & 0.01 & +0.53 & +0.01 & 59.87 & 0.01 & +0.50 & +0.02 & 55.38 & 0.01 & -0.13 & -0.00 & 69.94 \\
User / Location & Eastern Europe & 2.95\% & 0.00 & -0.50 & -0.01 & 32.11 & 0.00 & +0.07 & +0.00 & 33.73 & 0.00 & -0.18 & -0.01 & 30.27 \\
User / Location & Western Europe & 5.37\% & 0.00 & +0.06 & +0.00 & 38.54 & 0.00 & +0.18 & +0.01 & 18.69 & 0.00 & -0.03 & -0.00 & 38.73 \\
User / Location & Southern Europe & 4.16\% & 0.00 & +0.90 & +0.02 & 24.59 & 0.00 & +0.90 & +0.08 & 9.70 & 0.00 & -0.82 & -0.04 & 14.02 \\
User / Location & Prefer not to Say & 0.08\% & 0.00 & +0.99 & +0.19 & 2.69 & 0.00 & +0.96 & +0.29 & 1.44 & 0.00 & -0.91 & -0.13 & 4.22 \\

\SetHline{1-Z}{lightgray, 0.5pt, dashed}
User / Marital & Married & 30.04\% & 0.01 & -0.88 & -0.03 & 29.70 & 0.01 & -0.46 & -0.02 & 45.98 & 0.01 & +0.14 & +0.00 & 68.95 \\
User / Marital & Never Married & 59.45\% & 0.01 & -0.79 & -0.02 & 73.28 & 0.01 & -0.09 & -0.00 & 94.95 & 0.01 & -0.18 & -0.00 & 104.23 \\
User / Marital & Divorced & 7.71\% & 0.01 & -0.88 & -0.04 & 13.84 & 0.01 & -0.52 & -0.04 & 18.59 & 0.00 & +0.01 & +0.00 & 33.31 \\
User / Marital & Widowed & 1.47\% & 0.00 & +0.27 & +0.01 & 26.89 & 0.00 & -0.29 & -0.02 & 21.99 & 0.00 & +0.45 & +0.01 & 33.03 \\
User / Marital & Prefer not to Say & 1.33\% & 0.00 & -0.50 & -0.02 & 17.41 & 0.00 & -0.85 & -0.10 & 7.12 & 0.00 & +0.92 & +0.07 & 8.97 \\

\SetHline{1-Z}{lightgray, 0.5pt, dashed}
User / Religion & Agnostic & 4.87\% & 0.00 & -0.46 & -0.01 & 29.30 & 0.00 & -0.54 & -0.04 & 20.30 & 0.00 & +0.17 & +0.00 & 30.09 \\
User / Religion & Buddhist & 0.65\% & 0.00 & -0.11 & -0.00 & 24.63 & 0.00 & -0.86 & -0.11 & 7.60 & 0.00 & +0.91 & +0.10 & 6.03 \\
User / Religion & Christian & 31.81\% & 0.00 & -0.31 & -0.00 & 79.01 & 0.01 & -0.11 & -0.00 & 59.95 & 0.01 & -0.23 & -0.00 & 61.52 \\
User / Religion & Folk Religion & 0.46\% & 0.00 & -0.99 & -0.18 & 2.97 & 0.00 & -0.85 & -0.15 & 3.76 & 0.00 & +0.19 & +0.01 & 26.74 \\
User / Religion & Hindu & 0.33\% & 0.00 & -0.64 & -0.02 & 20.15 & 0.00 & -0.69 & -0.04 & 16.75 & 0.00 & +0.60 & +0.02 & 27.41 \\
User / Religion & Jewish & 2.59\% & 0.00 & +0.89 & +0.03 & 16.82 & 0.01 & +0.97 & +0.15 & 4.11 & 0.01 & -0.91 & -0.09 & 6.32 \\
User / Religion & Muslim & 2.20\% & 0.01 & -0.98 & -0.13 & 3.70 & 0.00 & -0.84 & -0.07 & 10.81 & 0.00 & -0.74 & -0.03 & 18.59 \\
User / Religion & Non-Religious & 51.61\% & 0.01 & +0.05 & +0.00 & 64.58 & 0.02 & -0.83 & -0.04 & 46.78 & 0.01 & +0.77 & +0.01 & 69.59 \\
User / Religion & Sikh & 0.23\% & 0.00 & -0.98 & -0.27 & 1.90 & 0.00 & -0.91 & -0.13 & 4.54 & 0.00 & -0.98 & -0.12 & 4.53 \\
User / Religion & Spiritual & 1.07\% & 0.00 & +0.98 & +0.12 & 4.17 & 0.00 & -0.08 & -0.01 & 14.83 & 0.00 & +0.83 & +0.06 & 8.67 \\
User / Religion & Other & 0.20\% & 0.00 & -0.95 & -0.24 & 2.17 & 0.00 & -0.86 & -0.15 & 4.88 & 0.00 & -0.78 & -0.06 & 10.39 \\
User / Religion & Prefer not to Say & 3.97\% & 0.00 & -0.92 & -0.05 & 10.22 & 0.00 & -0.09 & -0.00 & 37.34 & 0.00 & -0.63 & -0.02 & 20.22 \\

\SetHline{1-Z}{lightgray, 0.5pt, solid}
AnthroScore & Anthro Score & 29.04\% & 0.01 & +0.49 & +0.01 & 27.49 & 0.01 & +0.62 & +0.01 & 31.36 & 0.01 & -0.05 & -0.00 & 39.49 \\

\SetHline{1-Z}{lightgray, 0.5pt, solid}
Complexity & Num Lemmas & 97.76\% & 0.07 & +0.90 & +0.02 & 7.24 & 0.16 & +0.95 & +0.04 & 3.19 & 0.10 & -0.95 & -0.03 & 4.97 \\
Complexity & Num Tokens & 98.19\% & 0.04 & +0.69 & +0.01 & 12.96 & 0.07 & +0.86 & +0.02 & 10.61 & 0.05 & -0.78 & -0.01 & 11.64 \\
Complexity & Num Sents & 80.20\% & 0.02 & +0.12 & +0.00 & 26.71 & 0.07 & +0.82 & +0.02 & 9.04 & 0.04 & -0.76 & -0.01 & 12.79 \\
Complexity & Type-to-Token Ratio & 99.27\% & 0.03 & -0.99 & -0.26 & 20.28 & 0.06 & -0.99 & -0.52 & 15.71 & 0.04 & +0.98 & +0.31 & 19.16 \\
Complexity & Num Entities & 63.23\% & 0.03 & +0.21 & +0.00 & 14.76 & 0.03 & +0.54 & +0.01 & 22.25 & 0.02 & -0.22 & -0.00 & 25.73 \\
Complexity & Avg Sent Length & 98.77\% & 0.04 & -0.32 & -0.00 & 13.11 & 0.04 & +0.08 & +0.00 & 21.78 & 0.02 & -0.07 & -0.00 & 26.41 \\
Complexity & Syllables per Word & 99.03\% & 0.04 & +0.99 & +0.21 & 14.11 & 0.06 & +0.98 & +0.33 & 14.04 & 0.02 & -0.96 & -0.13 & 27.76 \\
Complexity & Entities per Sent & 68.81\% & 0.02 & +0.70 & +0.02 & 23.05 & 0.03 & +0.84 & +0.04 & 21.99 & 0.01 & -0.66 & -0.02 & 35.80 \\

\pagebreak

\SetHline{1-Z}{lightgray, 0.5pt, solid}
Dialog Acts & Comment & 94.63\% & 0.01 & +0.21 & +0.00 & 91.19 & 0.06 & +0.97 & +0.08 & 15.35 & 0.06 & -0.97 & -0.08 & 10.02 \\
Dialog Acts & Respond to Apology & 94.63\% & 0.02 & +0.93 & +0.03 & 36.97 & 0.06 & +0.97 & +0.09 & 12.74 & 0.05 & -0.98 & -0.08 & 11.08 \\
Dialog Acts & Abandon & 94.63\% & 0.01 & +0.74 & +0.02 & 42.21 & 0.06 & +0.96 & +0.09 & 15.66 & 0.04 & -0.92 & -0.06 & 16.09 \\
Dialog Acts & Appreciation & 94.63\% & 0.00 & +0.46 & +0.00 & 97.07 & 0.03 & -0.95 & -0.06 & 25.86 & 0.04 & +0.97 & +0.06 & 17.69 \\
Dialog Acts & Opinion & 94.63\% & 0.01 & +0.74 & +0.01 & 55.26 & 0.04 & +0.96 & +0.06 & 24.74 & 0.03 & -0.96 & -0.05 & 19.00 \\
Dialog Acts & Nonesense & 94.63\% & 0.02 & -0.95 & -0.03 & 28.26 & 0.05 & -0.96 & -0.07 & 20.50 & 0.03 & +0.95 & +0.04 & 20.51 \\
Dialog Acts & Question - Factual & 94.63\% & 0.02 & +0.93 & +0.03 & 29.13 & 0.05 & +0.96 & +0.07 & 17.10 & 0.03 & -0.94 & -0.04 & 25.34 \\
Dialog Acts & Complaint & 94.63\% & 0.02 & -0.92 & -0.03 & 28.66 & 0.04 & -0.90 & -0.06 & 23.35 & 0.03 & +0.92 & +0.04 & 26.54 \\
Dialog Acts & Apology & 94.63\% & 0.02 & -0.95 & -0.03 & 29.14 & 0.05 & -0.96 & -0.07 & 19.54 & 0.02 & +0.93 & +0.03 & 27.97 \\
Dialog Acts & Open Question - Opinion & 94.63\% & 0.02 & -0.95 & -0.03 & 25.48 & 0.05 & -0.97 & -0.07 & 16.70 & 0.02 & +0.92 & +0.03 & 29.15 \\
Dialog Acts & Statement & 94.63\% & 0.01 & -0.48 & -0.01 & 79.31 & 0.04 & -0.94 & -0.05 & 28.47 & 0.02 & +0.94 & +0.03 & 33.13 \\
Dialog Acts & Hold & 94.63\% & 0.01 & -0.80 & -0.01 & 53.22 & 0.03 & -0.93 & -0.05 & 25.11 & 0.02 & +0.89 & +0.03 & 35.19 \\
Dialog Acts & Answer - Pos & 94.63\% & 0.02 & -0.91 & -0.03 & 33.74 & 0.01 & +0.56 & +0.01 & 83.88 & 0.02 & -0.85 & -0.03 & 40.12 \\
Dialog Acts & Opening & 94.63\% & 0.05 & +0.99 & +0.08 & 9.71 & 0.06 & +0.97 & +0.10 & 12.52 & 0.02 & -0.90 & -0.02 & 43.54 \\
Dialog Acts & Other Answers & 94.63\% & 0.02 & -0.89 & -0.03 & 31.12 & 0.04 & -0.91 & -0.05 & 25.01 & 0.01 & +0.86 & +0.02 & 43.88 \\
Dialog Acts & Dev Command & 94.63\% & 0.02 & -0.87 & -0.02 & 35.85 & 0.03 & -0.84 & -0.04 & 33.00 & 0.01 & +0.77 & +0.02 & 54.59 \\
Dialog Acts & Command & 94.63\% & 0.03 & -0.95 & -0.04 & 19.95 & 0.03 & -0.88 & -0.05 & 29.83 & 0.01 & +0.67 & +0.01 & 57.37 \\
Dialog Acts & Back Channeling & 94.63\% & 0.01 & +0.46 & +0.01 & 70.63 & 0.01 & +0.64 & +0.02 & 66.20 & 0.01 & -0.67 & -0.01 & 57.50 \\
Dialog Acts & Other & 94.63\% & 0.04 & -0.97 & -0.05 & 16.45 & 0.02 & -0.87 & -0.03 & 49.20 & 0.01 & -0.57 & -0.01 & 59.67 \\
Dialog Acts & Answer - Neg & 94.63\% & 0.04 & -0.98 & -0.06 & 14.09 & 0.05 & -0.93 & -0.08 & 15.87 & 0.01 & +0.57 & +0.01 & 67.20 \\
Dialog Acts & Question - Yes / No & 94.63\% & 0.01 & -0.10 & -0.00 & 84.92 & 0.01 & -0.09 & -0.00 & 96.00 & 0.01 & +0.43 & +0.01 & 78.87 \\
Dialog Acts & Closing Delta & 94.63\% & 0.01 & +0.42 & +0.00 & 79.67 & 0.01 & +0.50 & +0.01 & 69.68 & 0.01 & +0.02 & +0.00 & 83.45 \\
Dialog Acts & Thanking & 94.63\% & 0.01 & -0.75 & -0.01 & 60.49 & 0.01 & -0.08 & -0.00 & 93.59 & 0.01 & +0.38 & +0.00 & 88.08 \\

\SetHline{1-Z}{lightgray, 0.5pt, solid}
Emotions & Approval & 100.00\% & 0.01 & +0.54 & +0.01 & 46.37 & 0.04 & +0.91 & +0.06 & 19.70 & 0.04 & -0.94 & -0.05 & 17.10 \\
Emotions & Anger & 100.00\% & 0.02 & -0.90 & -0.03 & 30.77 & 0.05 & -0.93 & -0.06 & 17.88 & 0.03 & +0.92 & +0.04 & 24.31 \\
Emotions & Caring & 100.00\% & 0.01 & +0.34 & +0.01 & 44.07 & 0.02 & +0.82 & +0.03 & 38.58 & 0.02 & -0.73 & -0.03 & 28.23 \\
Emotions & Nervousness & 100.00\% & 0.01 & +0.60 & +0.01 & 62.63 & 0.01 & -0.77 & -0.02 & 70.11 & 0.02 & +0.87 & +0.03 & 31.24 \\
Emotions & Pride & 100.00\% & 0.01 & -0.74 & -0.02 & 42.32 & 0.04 & -0.85 & -0.05 & 23.85 & 0.02 & +0.87 & +0.03 & 32.83 \\
Emotions & Love & 100.00\% & 0.01 & +0.79 & +0.01 & 46.42 & 0.03 & +0.89 & +0.04 & 27.40 & 0.02 & -0.89 & -0.02 & 34.72 \\
Emotions & Grief & 100.00\% & 0.01 & +0.42 & +0.01 & 54.68 & 0.01 & -0.50 & -0.01 & 67.45 & 0.02 & +0.77 & +0.02 & 38.17 \\
Emotions & Disgust & 100.00\% & 0.01 & +0.13 & +0.00 & 59.23 & 0.02 & +0.76 & +0.03 & 40.57 & 0.01 & -0.82 & -0.02 & 44.11 \\
Emotions & Fear & 100.00\% & 0.01 & +0.20 & +0.00 & 82.05 & 0.02 & +0.61 & +0.02 & 51.99 & 0.01 & -0.87 & -0.02 & 44.18 \\
Emotions & Optimism & 100.00\% & 0.01 & -0.71 & -0.01 & 55.17 & 0.02 & -0.78 & -0.03 & 37.74 & 0.01 & +0.76 & +0.02 & 45.59 \\
Emotions & Desire & 100.00\% & 0.01 & -0.27 & -0.00 & 71.11 & 0.01 & -0.54 & -0.01 & 66.65 & 0.01 & +0.84 & +0.02 & 48.53 \\
Emotions & Surprise & 100.00\% & 0.01 & +0.59 & +0.01 & 48.84 & 0.01 & +0.33 & +0.01 & 64.22 & 0.01 & -0.33 & -0.01 & 51.39 \\
Emotions & Neutral & 100.00\% & 0.03 & +0.90 & +0.04 & 21.38 & 0.04 & +0.91 & +0.05 & 24.43 & 0.01 & -0.75 & -0.01 & 52.71 \\
Emotions & Realization & 100.00\% & 0.01 & +0.13 & +0.00 & 59.35 & 0.01 & -0.06 & -0.00 & 60.27 & 0.01 & +0.34 & +0.01 & 53.81 \\
Emotions & Disapproval & 100.00\% & 0.01 & +0.81 & +0.02 & 45.38 & 0.03 & +0.89 & +0.04 & 33.53 & 0.01 & -0.74 & -0.01 & 54.28 \\
Emotions & Remorse & 100.00\% & 0.01 & -0.86 & -0.02 & 41.69 & 0.02 & -0.88 & -0.04 & 35.30 & 0.01 & +0.65 & +0.02 & 54.69 \\
Emotions & Confusion & 100.00\% & 0.01 & -0.78 & -0.02 & 40.32 & 0.02 & +0.31 & +0.01 & 56.20 & 0.01 & -0.33 & -0.01 & 56.74 \\
Emotions & Amusement & 100.00\% & 0.02 & +0.72 & +0.02 & 33.36 & 0.02 & +0.67 & +0.02 & 37.75 & 0.01 & -0.09 & -0.00 & 58.69 \\
Emotions & Relief & 100.00\% & 0.01 & -0.35 & -0.01 & 60.18 & 0.02 & +0.34 & +0.01 & 51.45 & 0.01 & -0.68 & -0.01 & 59.15 \\
Emotions & Disappointment & 100.00\% & 0.01 & -0.71 & -0.01 & 54.97 & 0.01 & -0.50 & -0.01 & 61.75 & 0.01 & +0.13 & +0.00 & 59.31 \\
Emotions & Sadness & 100.00\% & 0.01 & +0.21 & +0.00 & 70.06 & 0.01 & +0.58 & +0.01 & 61.84 & 0.01 & -0.18 & -0.00 & 59.98 \\
Emotions & Gratitude & 100.00\% & 0.01 & -0.70 & -0.01 & 48.79 & 0.01 & -0.15 & -0.00 & 69.16 & 0.01 & -0.45 & -0.01 & 66.18 \\
Emotions & Excitement & 100.00\% & 0.01 & -0.28 & -0.00 & 90.83 & 0.01 & +0.24 & +0.01 & 64.68 & 0.01 & -0.33 & -0.01 & 66.67 \\
Emotions & Annoyance & 100.00\% & 0.01 & -0.29 & -0.00 & 65.53 & 0.01 & -0.28 & -0.01 & 74.76 & 0.01 & -0.18 & -0.00 & 70.88 \\
Emotions & Admiration & 100.00\% & 0.01 & +0.12 & +0.00 & 43.86 & 0.01 & -0.18 & -0.00 & 62.65 & 0.01 & +0.28 & +0.00 & 71.50 \\
Emotions & Curiosity & 100.00\% & 0.02 & +0.79 & +0.02 & 30.86 & 0.02 & +0.61 & +0.02 & 55.38 & 0.01 & +0.18 & +0.00 & 72.43 \\
Emotions & Joy & 100.00\% & 0.01 & -0.10 & -0.00 & 69.57 & 0.01 & -0.38 & -0.01 & 64.82 & 0.01 & +0.45 & +0.01 & 73.46 \\
Emotions & Embarrassment & 100.00\% & 0.01 & +0.71 & +0.01 & 46.01 & 0.02 & +0.63 & +0.02 & 49.12 & 0.01 & -0.13 & -0.00 & 74.48 \\

\SetHline{1-Z}{lightgray, 0.5pt, solid}
Politeness & Impolite & 99.99\% & 0.01 & -0.42 & -0.01 & 69.84 & 0.03 & -0.90 & -0.05 & 27.12 & 0.03 & +0.94 & +0.04 & 20.77 \\
Politeness & Neutral & 99.99\% & 0.01 & +0.54 & +0.01 & 57.72 & 0.02 & -0.66 & -0.02 & 50.44 & 0.02 & +0.89 & +0.03 & 30.20 \\
Politeness & Somewhat Polite & 99.99\% & 0.02 & +0.79 & +0.02 & 36.62 & 0.01 & -0.01 & -0.00 & 78.10 & 0.01 & +0.78 & +0.02 & 44.21 \\
Politeness & Polite & 99.99\% & 0.02 & +0.81 & +0.02 & 37.39 & 0.01 & +0.45 & +0.01 & 68.11 & 0.02 & +0.84 & +0.02 & 42.02 \\

\pagebreak
\SetHline{1-Z}{lightgray, 0.5pt, solid}
Rule-based Rewards & Partial Compliance & 23.23\% & 0.05 & +0.96 & +0.14 & 4.76 & 0.14 & +0.98 & +0.39 & 1.64 & 0.09 & -0.97 & -0.25 & 2.27 \\
Rule-based Rewards & Third Person & 0.93\% & 0.00 & -0.45 & -0.02 & 15.58 & 0.00 & -0.96 & -0.27 & 2.55 & 0.00 & +0.97 & +0.21 & 2.56 \\
Rule-based Rewards & Full Compliance & 35.24\% & 0.07 & +0.98 & +0.18 & 2.96 & 0.13 & +0.99 & +0.30 & 2.02 & 0.05 & -0.96 & -0.12 & 5.08 \\
Rule-based Rewards & Disclaimer & 2.11\% & 0.01 & -0.95 & -0.12 & 4.38 & 0.01 & -0.97 & -0.21 & 2.99 & 0.00 & +0.95 & +0.11 & 5.32 \\
Rule-based Rewards & Meta Commentary & 11.88\% & 0.01 & -0.95 & -0.11 & 5.26 & 0.03 & -0.98 & -0.22 & 2.96 & 0.01 & +0.96 & +0.11 & 5.34 \\
Rule-based Rewards & Provides Resources & 1.13\% & 0.00 & +0.21 & +0.01 & 20.38 & 0.00 & +0.89 & +0.09 & 9.23 & 0.00 & -0.92 & -0.07 & 7.27 \\
Rule-based Rewards & Judgement & 3.38\% & 0.00 & +0.98 & +0.08 & 7.53 & 0.00 & +0.95 & +0.09 & 11.58 & 0.00 & -0.86 & -0.04 & 17.74 \\
Rule-based Rewards & Prescribes Solutions & 5.65\% & 0.00 & +0.37 & +0.01 & 32.67 & 0.00 & +0.16 & +0.00 & 52.30 & 0.00 & -0.79 & -0.02 & 25.14 \\
Rule-based Rewards & Definitive Verbiage & 1.12\% & 0.00 & -0.83 & -0.02 & 28.50 & 0.00 & -0.77 & -0.03 & 27.68 & 0.00 & -0.15 & -0.00 & 30.29 \\
Rule-based Rewards & Non-Compliance & 1.65\% & 0.00 & -0.97 & -0.07 & 9.11 & 0.00 & -0.96 & -0.12 & 8.21 & 0.00 & +0.67 & +0.02 & 31.14 \\
Rule-based Rewards & Gentle Encouragement for Help & 2.12\% & 0.00 & -0.50 & -0.01 & 25.14 & 0.00 & -0.63 & -0.03 & 23.26 & 0.00 & +0.54 & +0.01 & 41.80 \\
Rule-based Rewards & Safety Policy & 1.04\% & 0.00 & +0.95 & +0.08 & 6.79 & 0.00 & +0.65 & +0.03 & 18.14 & 0.00 & +0.08 & +0.00 & 43.55 \\
Rule-based Rewards & Hedging & 28.77\% & 0.01 & -0.85 & -0.02 & 28.12 & 0.01 & +0.29 & +0.01 & 54.63 & 0.00 & -0.69 & -0.01 & 43.72 \\
Rule-based Rewards & Professional Help & 5.11\% & 0.00 & -0.73 & -0.02 & 25.45 & 0.00 & -0.27 & -0.01 & 40.50 & 0.00 & -0.10 & -0.00 & 50.81 \\
Rule-based Rewards & Acknowledges Emotional State & 0.93\% & 0.00 & -0.59 & -0.02 & 25.98 & 0.00 & +0.38 & +0.02 & 27.70 & 0.00 & +0.34 & +0.01 & 57.17 \\

\SetHline{1-Z}{lightgray, 0.5pt, solid}
Sycophancy & Framing & 14.53\% & 0.01 & -0.96 & -0.07 & 8.89 & 0.03 & -0.98 & -0.18 & 4.41 & 0.02 & +0.97 & +0.12 & 5.35 \\
Sycophancy & Validation & 8.13\% & 0.01 & +0.91 & +0.05 & 12.14 & 0.01 & +0.94 & +0.09 & 9.06 & 0.01 & -0.94 & -0.05 & 13.71 \\
Sycophancy & Indirectness & 27.41\% & 0.00 & +0.64 & +0.01 & 42.46 & 0.02 & +0.86 & +0.04 & 22.72 & 0.01 & -0.78 & -0.03 & 23.82 \\

\end{longtblr}

\begin{longtblr}[
    caption = {Summary statistics for individual SHAP values on \community.},
    label = {tab:regression_community_individual_values},
]{
    width = 1.0\textwidth,
    colspec = {X[c]X[l]cccccccccccc},
    colsep = 2.0pt,
    rowhead = 2,
    hline{1, 3, Z} = {solid},
    cells = {
        font=\footnotesize
    },
    row{1} = {font=\bfseries},
    row{2} = {font=\itshape\small},
    cell{1}{1} = {r=2,c=1}{c},
    cell{1}{2} = {r=2,c=1}{c},
    cell{1}{3} = {r=2,c=1}{c},
    cell{1}{4} = {r=1,c=4}{c},
    cell{1}{8} = {r=1,c=4}{c},
    cell{1}{12} = {r=1,c=4}{c},
    row{3-Z} = {font = \tiny, abovesep = -3pt, belowsep = -3pt},
    cell{3}{1} = {r=1,c=2}{c},
    cell{4}{1} = {r=1,c=2}{c},
    cell{5}{1} = {r=4,c=1}{c},
    cell{9}{1} = {r=2,c=1}{c},
    cell{11}{1} = {r=5,c=1}{c},
    cell{16}{1} = {r=7,c=1}{c},
    cell{23}{1} = {r=8,c=1}{c},
    cell{31}{1} = {r=2,c=1}{c},
    cell{33}{1} = {r=1,c=1}{c},
    cell{34}{1} = {r=8,c=1}{c},
    cell{42}{1} = {r=23,c=1}{c},
    cell{65}{1} = {r=28,c=1}{c},
    cell{93}{1} = {r=4,c=1}{c},
    cell{97}{1} = {r=15,c=1}{c},
    cell{112}{1} = {r=3,c=1}{c},
}

Block & Name & Prop & \tm & & & & \tg & & & & \cm \\
                & & & MAV & r & $\beta$ & HMR & MAV & r & $\beta$ & HMR & MAV & r & $\beta$ & HMR \\

Turn Entropy &  & 27.24\% & 0.41 & -0.98 & -1.85 & 1.26 & 0.16 & -0.91 & -0.71 & 2.08 & 0.18 & -0.93 & -0.83 & 1.63 \\

\SetHline{1-Z}{lightgray, 0.5pt, solid}
Margin &  & 98.74\% & 0.45 & +0.95 & +2.28 & 1.91 & 0.26 & +0.79 & +1.29 & 4.13 & 0.31 & +0.94 & +1.17 & 1.42 \\

\SetHline{1-Z}{lightgray, 0.5pt, solid}
User / Age & 18-34 & 46.38\% & 0.02 & -0.62 & -0.03 & 31.35 & 0.03 & +0.31 & +0.03 & 24.80 & 0.01 & +0.12 & +0.00 & 40.78 \\
User / Age & 35-45 & 21.09\% & 0.02 & -0.88 & -0.05 & 11.14 & 0.03 & -0.73 & -0.08 & 9.80 & 0.01 & +0.60 & +0.02 & 23.97 \\
User / Age & 46-54 & 16.79\% & 0.03 & -0.93 & -0.08 & 6.99 & 0.05 & -0.89 & -0.16 & 4.65 & 0.01 & +0.81 & +0.04 & 15.65 \\
User / Age & 55+ & 14.46\% & 0.03 & -0.82 & -0.10 & 6.13 & 0.06 & -0.86 & -0.22 & 3.20 & 0.02 & +0.83 & +0.06 & 11.86 \\

\SetHline{1-Z}{lightgray, 0.5pt, dashed}
User / Gender & female & 50.99\% & 0.05 & -0.96 & -0.10 & 9.42 & 0.04 & -0.68 & -0.06 & 20.73 & 0.02 & -0.76 & -0.03 & 30.78 \\
User / Gender & male & 48.80\% & 0.05 & -0.96 & -0.09 & 11.35 & 0.04 & +0.11 & +0.01 & 26.95 & 0.04 & -0.88 & -0.07 & 18.83 \\

\SetHline{1-Z}{lightgray, 0.5pt, dashed}
User / Education & (At most) Complete Secondary & 22.28\% & 0.22 & -0.96 & -0.59 & 1.14 & 0.31 & -0.93 & -0.82 & 1.06 & 0.13 & +0.94 & +0.34 & 1.24 \\
User / Education & Some post-secondary & 6.56\% & 0.03 & -0.87 & -0.22 & 2.63 & 0.03 & -0.78 & -0.24 & 2.37 & 0.01 & +0.60 & +0.06 & 7.10 \\
User / Education & Post-secondary graduate & 36.84\% & 0.02 & +0.64 & +0.04 & 16.22 & 0.05 & +0.51 & +0.07 & 12.38 & 0.03 & -0.86 & -0.07 & 12.48 \\
User / Education & Some or complete graduate degree & 33.29\% & 0.02 & +0.17 & +0.01 & 26.37 & 0.05 & +0.56 & +0.07 & 13.94 & 0.04 & -0.90 & -0.08 & 14.06 \\
User / Education & Other & 0.18\% & 0.00 & -0.03 & -0.00 & 16.41 & 0.00 & -0.31 & -0.03 & 11.81 & 0.00 & +0.43 & +0.02 & 15.62 \\

\SetHline{1-Z}{lightgray, 0.5pt, dashed}
User / Political & Very left-leaning & 7.96\% & 0.01 & -0.70 & -0.04 & 12.83 & 0.01 & -0.14 & -0.01 & 12.12 & 0.01 & -0.67 & -0.04 & 12.76 \\
User / Political & Somewhat left-leaning & 19.65\% & 0.01 & -0.54 & -0.02 & 23.52 & 0.02 & +0.52 & +0.05 & 12.22 & 0.02 & -0.80 & -0.07 & 9.53 \\
User / Political & Middle-of-the-road, centrist & 26.31\% & 0.01 & -0.68 & -0.02 & 29.40 & 0.03 & +0.64 & +0.07 & 14.07 & 0.03 & -0.85 & -0.07 & 11.76 \\
User / Political & Somewhat right-leaning & 9.78\% & 0.01 & +0.21 & +0.01 & 14.02 & 0.03 & +0.64 & +0.10 & 6.37 & 0.02 & -0.91 & -0.10 & 6.49 \\
User / Political & Very right-leaning & 6.81\% & 0.00 & -0.34 & -0.01 & 21.52 & 0.01 & +0.76 & +0.10 & 5.88 & 0.01 & -0.93 & -0.08 & 6.50 \\
User / Political & Prefer not to say & 7.74\% & 0.01 & -0.28 & -0.01 & 12.87 & 0.01 & +0.19 & +0.02 & 12.37 & 0.00 & -0.61 & -0.03 & 17.90 \\
User / Political & I don't think of myself in this way & 21.75\% & 0.02 & -0.73 & -0.04 & 17.00 & 0.03 & +0.60 & +0.06 & 11.78 & 0.03 & -0.87 & -0.09 & 7.96 \\

\SetHline{1-Z}{lightgray, 0.5pt, dashed}
User / Ethnicity & Asian & 3.09\% & 0.00 & +0.82 & +0.06 & 8.28 & 0.01 & +0.87 & +0.22 & 2.05 & 0.01 & -0.93 & -0.13 & 3.03 \\
User / Ethnicity & Black or African American & 4.10\% & 0.01 & -0.88 & -0.07 & 7.76 & 0.01 & -0.40 & -0.04 & 10.69 & 0.01 & -0.72 & -0.08 & 5.09 \\
User / Ethnicity & Dravidian & 14.05\% & 0.02 & -0.94 & -0.08 & 6.75 & 0.02 & -0.63 & -0.07 & 8.18 & 0.01 & -0.83 & -0.04 & 14.70 \\
User / Ethnicity & Hispanic or Latino & 6.56\% & 0.01 & -0.87 & -0.09 & 6.12 & 0.01 & -0.65 & -0.09 & 5.30 & 0.01 & -0.45 & -0.03 & 13.63 \\
User / Ethnicity & Indo-Aryan & 38.92\% & 0.04 & -0.95 & -0.09 & 10.02 & 0.04 & -0.41 & -0.04 & 17.28 & 0.03 & -0.86 & -0.06 & 16.58 \\
User / Ethnicity & Other & 3.27\% & 0.01 & +0.67 & +0.09 & 5.42 & 0.02 & +0.76 & +0.29 & 1.77 & 0.01 & -0.83 & -0.09 & 5.06 \\
User / Ethnicity & Prefer not to say & 5.17\% & 0.00 & +0.39 & +0.01 & 19.02 & 0.01 & +0.48 & +0.05 & 9.50 & 0.01 & -0.87 & -0.12 & 3.18 \\
User / Ethnicity & White & 24.84\% & 0.02 & +0.51 & +0.04 & 13.29 & 0.05 & +0.82 & +0.16 & 4.54 & 0.04 & -0.92 & -0.12 & 6.07 \\

\SetHline{1-Z}{lightgray, 0.5pt, dashed}
User / Country & india & 55.80\% & 0.04 & -0.86 & -0.08 & 18.15 & 0.05 & +0.61 & +0.08 & 20.70 & 0.09 & -0.95 & -0.19 & 6.50 \\
User / Country & united states & 44.20\% & 0.04 & -0.90 & -0.09 & 9.33 & 0.06 & +0.57 & +0.08 & 12.10 & 0.08 & -0.95 & -0.17 & 5.37 \\

\SetHline{1-Z}{lightgray, 0.5pt, solid}
AnthroScore & Anthro Score & 15.37\% & 0.00 & -0.46 & -0.01 & 31.19 & 0.01 & -0.63 & -0.04 & 15.39 & 0.01 & +0.72 & +0.04 & 12.05 \\

\SetHline{1-Z}{lightgray, 0.5pt, solid}
Complexity & Num Tokens & 97.69\% & 0.03 & +0.03 & +0.00 & 17.63 & 0.07 & +0.43 & +0.01 & 10.31 & 0.05 & -0.23 & -0.00 & 11.16 \\
Complexity & Num Sents & 59.75\% & 0.03 & -0.70 & -0.01 & 13.75 & 0.06 & -0.30 & -0.01 & 8.22 & 0.02 & +0.25 & +0.00 & 16.64 \\
Complexity & Num Lemmas & 96.09\% & 0.03 & -0.63 & -0.01 & 17.85 & 0.05 & -0.24 & -0.00 & 17.25 & 0.03 & +0.07 & +0.00 & 18.53 \\
Complexity & Avg Sent Length & 98.30\% & 0.05 & -0.59 & -0.00 & 10.52 & 0.09 & -0.58 & -0.01 & 7.04 & 0.04 & +0.10 & +0.00 & 13.27 \\
Complexity & Syllables per Word & 99.79\% & 0.02 & -0.83 & -0.13 & 28.63 & 0.04 & -0.83 & -0.32 & 20.59 & 0.02 & +0.86 & +0.21 & 23.01 \\
Complexity & Type-to-Token Ratio & 99.54\% & 0.01 & -0.78 & -0.11 & 58.50 & 0.01 & -0.82 & -0.23 & 55.84 & 0.01 & +0.75 & +0.15 & 58.75 \\
Complexity & Num Entities & 69.42\% & 0.04 & +0.54 & +0.01 & 13.35 & 0.07 & +0.31 & +0.01 & 9.13 & 0.03 & -0.22 & -0.00 & 13.83 \\
Complexity & Entities per Sent & 75.98\% & 0.02 & +0.68 & +0.02 & 26.17 & 0.03 & +0.50 & +0.03 & 23.17 & 0.01 & +0.07 & +0.00 & 34.72 \\

\SetHline{1-Z}{lightgray, 0.5pt, solid}
Dialog Acts & Question - Factual & 90.07\% & 0.02 & -0.61 & -0.01 & 34.06 & 0.03 & -0.58 & -0.03 & 26.75 & 0.01 & +0.80 & +0.02 & 40.01 \\
Dialog Acts & Thanking & 90.07\% & 0.02 & -0.21 & -0.01 & 32.30 & 0.05 & -0.76 & -0.07 & 19.04 & 0.05 & +0.85 & +0.07 & 12.13 \\
Dialog Acts & Hold & 90.07\% & 0.01 & +0.13 & +0.00 & 42.44 & 0.03 & +0.78 & +0.04 & 33.76 & 0.02 & -0.88 & -0.03 & 28.40 \\
Dialog Acts & Respond to Apology & 90.07\% & 0.02 & +0.84 & +0.03 & 34.71 & 0.01 & +0.49 & +0.01 & 55.49 & 0.01 & -0.60 & -0.01 & 57.13 \\
Dialog Acts & Apology & 90.07\% & 0.02 & -0.84 & -0.03 & 25.71 & 0.02 & -0.18 & -0.00 & 51.45 & 0.01 & -0.48 & -0.01 & 55.26 \\
Dialog Acts & Complaint & 90.07\% & 0.03 & -0.57 & -0.02 & 21.02 & 0.03 & -0.49 & -0.02 & 29.50 & 0.01 & +0.55 & +0.01 & 40.13 \\
Dialog Acts & Statement & 90.07\% & 0.01 & +0.65 & +0.02 & 35.37 & 0.03 & +0.90 & +0.04 & 33.28 & 0.06 & -0.92 & -0.08 & 10.08 \\
Dialog Acts & Back Channeling & 90.07\% & 0.01 & -0.31 & -0.01 & 53.48 & 0.02 & +0.23 & +0.01 & 43.41 & 0.02 & -0.87 & -0.03 & 32.18 \\
Dialog Acts & Open Question - Opinion & 90.07\% & 0.03 & +0.87 & +0.05 & 16.20 & 0.06 & +0.84 & +0.09 & 12.24 & 0.03 & -0.79 & -0.04 & 19.61 \\
Dialog Acts & Other Answers & 90.07\% & 0.04 & -0.93 & -0.06 & 15.17 & 0.09 & -0.92 & -0.12 & 8.40 & 0.05 & +0.92 & +0.08 & 11.51 \\
Dialog Acts & Closing Delta & 90.07\% & 0.02 & +0.83 & +0.03 & 25.23 & 0.04 & +0.78 & +0.05 & 23.22 & 0.02 & -0.58 & -0.02 & 36.09 \\
Dialog Acts & Nonesense & 90.07\% & 0.04 & -0.96 & -0.06 & 13.42 & 0.08 & -0.88 & -0.10 & 8.36 & 0.07 & +0.91 & +0.09 & 6.14 \\
Dialog Acts & Question - Yes / No & 90.07\% & 0.01 & +0.33 & +0.01 & 34.39 & 0.03 & +0.23 & +0.01 & 27.37 & 0.02 & +0.74 & +0.03 & 27.55 \\
Dialog Acts & Appreciation & 90.07\% & 0.03 & -0.80 & -0.04 & 17.67 & 0.07 & -0.89 & -0.10 & 10.47 & 0.03 & +0.84 & +0.04 & 26.06 \\
Dialog Acts & Opening & 90.07\% & 0.02 & -0.94 & -0.03 & 27.17 & 0.03 & +0.46 & +0.02 & 35.38 & 0.03 & -0.91 & -0.04 & 26.42 \\
Dialog Acts & Dev Command & 90.07\% & 0.01 & +0.24 & +0.01 & 45.15 & 0.02 & +0.06 & +0.00 & 47.88 & 0.01 & -0.15 & -0.00 & 52.56 \\
Dialog Acts & Other & 90.07\% & 0.04 & -0.93 & -0.05 & 15.30 & 0.06 & -0.86 & -0.08 & 12.01 & 0.04 & +0.86 & +0.05 & 17.34 \\
Dialog Acts & Abandon & 90.07\% & 0.01 & +0.35 & +0.01 & 43.90 & 0.02 & -0.59 & -0.03 & 33.46 & 0.02 & +0.77 & +0.03 & 25.83 \\
Dialog Acts & Command & 90.07\% & 0.01 & -0.53 & -0.01 & 40.17 & 0.03 & -0.59 & -0.03 & 29.96 & 0.03 & +0.82 & +0.04 & 19.36 \\
Dialog Acts & Answer - Pos & 90.07\% & 0.02 & +0.78 & +0.02 & 29.84 & 0.02 & +0.29 & +0.01 & 40.86 & 0.01 & +0.19 & +0.00 & 53.85 \\
Dialog Acts & Answer - Neg & 90.07\% & 0.01 & -0.23 & -0.00 & 41.21 & 0.03 & -0.38 & -0.02 & 33.30 & 0.01 & +0.36 & +0.01 & 46.43 \\
Dialog Acts & Opinion & 90.07\% & 0.05 & -0.92 & -0.07 & 11.44 & 0.06 & -0.81 & -0.09 & 14.82 & 0.05 & +0.87 & +0.08 & 10.14 \\
Dialog Acts & Comment & 90.07\% & 0.03 & +0.81 & +0.04 & 22.53 & 0.03 & +0.73 & +0.05 & 25.23 & 0.01 & -0.22 & -0.01 & 41.27 \\

\SetHline{1-Z}{lightgray, 0.5pt, solid}
Emotions & Joy & 99.96\% & 0.01 & -0.04 & -0.00 & 49.04 & 0.01 & -0.40 & -0.01 & 46.32 & 0.01 & +0.13 & +0.00 & 44.50 \\
Emotions & Desire & 99.96\% & 0.01 & +0.01 & +0.00 & 50.13 & 0.02 & -0.03 & -0.00 & 33.37 & 0.02 & +0.03 & +0.00 & 26.22 \\
Emotions & Amusement & 99.96\% & 0.01 & +0.63 & +0.01 & 40.75 & 0.03 & +0.69 & +0.03 & 27.93 & 0.02 & -0.45 & -0.01 & 32.16 \\
Emotions & Realization & 99.96\% & 0.01 & -0.07 & -0.00 & 48.48 & 0.02 & -0.21 & -0.01 & 46.10 & 0.01 & -0.08 & -0.00 & 45.09 \\
Emotions & Grief & 99.96\% & 0.01 & +0.31 & +0.01 & 36.91 & 0.02 & -0.24 & -0.01 & 39.98 & 0.01 & -0.11 & -0.00 & 39.23 \\
Emotions & Anger & 99.96\% & 0.01 & +0.69 & +0.01 & 41.62 & 0.02 & +0.45 & +0.02 & 37.10 & 0.01 & +0.03 & +0.00 & 38.48 \\
Emotions & Caring & 99.96\% & 0.01 & +0.03 & +0.00 & 40.04 & 0.02 & -0.05 & -0.00 & 40.69 & 0.01 & -0.29 & -0.01 & 34.72 \\
Emotions & Embarrassment & 99.96\% & 0.01 & +0.26 & +0.01 & 43.04 & 0.02 & +0.24 & +0.01 & 37.29 & 0.01 & -0.22 & -0.00 & 43.08 \\
Emotions & Annoyance & 99.96\% & 0.01 & -0.20 & -0.00 & 43.03 & 0.02 & +0.43 & +0.01 & 44.79 & 0.01 & -0.59 & -0.01 & 43.13 \\
Emotions & Remorse & 99.96\% & 0.02 & -0.67 & -0.02 & 33.25 & 0.02 & -0.03 & -0.00 & 38.36 & 0.01 & -0.12 & -0.00 & 42.87 \\
Emotions & Confusion & 99.96\% & 0.01 & -0.08 & -0.00 & 39.76 & 0.02 & +0.07 & +0.00 & 41.78 & 0.01 & -0.45 & -0.01 & 42.68 \\
Emotions & Gratitude & 99.96\% & 0.01 & +0.39 & +0.01 & 44.01 & 0.02 & +0.66 & +0.03 & 33.38 & 0.01 & -0.62 & -0.02 & 38.07 \\
Emotions & Fear & 99.96\% & 0.01 & +0.07 & +0.00 & 42.75 & 0.02 & +0.38 & +0.01 & 43.06 & 0.01 & -0.51 & -0.01 & 39.28 \\
Emotions & Sadness & 99.96\% & 0.01 & -0.09 & -0.00 & 41.20 & 0.02 & -0.24 & -0.01 & 37.62 & 0.01 & +0.22 & +0.00 & 37.10 \\
Emotions & Disapproval & 99.96\% & 0.01 & -0.60 & -0.01 & 36.74 & 0.03 & -0.72 & -0.04 & 27.44 & 0.01 & +0.63 & +0.02 & 36.44 \\
Emotions & Love & 99.96\% & 0.01 & -0.54 & -0.01 & 41.41 & 0.03 & -0.13 & -0.01 & 28.72 & 0.02 & +0.18 & +0.01 & 28.49 \\
Emotions & Curiosity & 99.96\% & 0.01 & +0.35 & +0.01 & 48.02 & 0.02 & +0.12 & +0.00 & 36.51 & 0.01 & -0.14 & -0.00 & 40.70 \\
Emotions & Approval & 99.96\% & 0.02 & -0.68 & -0.02 & 33.63 & 0.03 & -0.71 & -0.04 & 24.45 & 0.02 & +0.61 & +0.02 & 27.15 \\
Emotions & Excitement & 99.96\% & 0.03 & +0.69 & +0.03 & 18.63 & 0.04 & +0.78 & +0.05 & 19.12 & 0.02 & -0.68 & -0.03 & 26.57 \\
Emotions & Surprise & 99.96\% & 0.01 & -0.40 & -0.01 & 38.30 & 0.02 & -0.50 & -0.02 & 33.68 & 0.02 & +0.61 & +0.02 & 27.50 \\
Emotions & Neutral & 99.96\% & 0.01 & -0.59 & -0.02 & 35.73 & 0.04 & -0.55 & -0.04 & 24.48 & 0.03 & +0.68 & +0.03 & 23.29 \\
Emotions & Relief & 99.96\% & 0.02 & -0.76 & -0.02 & 32.61 & 0.02 & -0.18 & -0.01 & 35.48 & 0.01 & +0.52 & +0.01 & 37.62 \\
Emotions & Disappointment & 99.96\% & 0.01 & -0.51 & -0.01 & 38.21 & 0.02 & -0.04 & -0.00 & 33.86 & 0.01 & +0.04 & +0.00 & 35.16 \\
Emotions & Nervousness & 99.96\% & 0.01 & +0.59 & +0.01 & 36.37 & 0.02 & -0.28 & -0.01 & 39.12 & 0.01 & +0.36 & +0.01 & 42.04 \\
Emotions & Pride & 99.96\% & 0.01 & -0.71 & -0.01 & 39.58 & 0.02 & -0.48 & -0.02 & 36.06 & 0.02 & +0.17 & +0.00 & 34.89 \\
Emotions & Admiration & 99.96\% & 0.01 & -0.54 & -0.01 & 39.17 & 0.02 & -0.66 & -0.02 & 36.93 & 0.01 & +0.73 & +0.02 & 40.27 \\
Emotions & Disgust & 99.96\% & 0.01 & +0.21 & +0.00 & 42.47 & 0.02 & -0.04 & -0.00 & 41.75 & 0.01 & -0.18 & -0.00 & 46.05 \\
Emotions & Optimism & 99.96\% & 0.01 & -0.74 & -0.02 & 37.95 & 0.03 & -0.72 & -0.03 & 27.78 & 0.02 & +0.59 & +0.01 & 36.50 \\

\SetHline{1-Z}{lightgray, 0.5pt, solid}
Politeness & Impolite & 99.96\% & 0.02 & +0.29 & +0.01 & 27.97 & 0.03 & +0.15 & +0.01 & 20.95 & 0.01 & -0.12 & -0.00 & 29.17 \\
Politeness & Neutral & 99.96\% & 0.03 & +0.90 & +0.04 & 17.00 & 0.06 & +0.88 & +0.08 & 11.90 & 0.04 & -0.89 & -0.05 & 14.50 \\
Politeness & Somewhat Polite & 99.96\% & 0.01 & -0.31 & -0.01 & 41.88 & 0.03 & -0.50 & -0.02 & 26.75 & 0.02 & +0.37 & +0.01 & 22.61 \\
Politeness & Polite & 99.96\% & 0.05 & +0.91 & +0.06 & 11.41 & 0.07 & +0.90 & +0.10 & 9.12 & 0.03 & -0.88 & -0.05 & 17.43 \\

\pagebreak

\SetHline{1-Z}{lightgray, 0.5pt, solid}
Rule-based Rewards & Acknowledges Emotional State & 0.21\% & 0.00 & +0.82 & +0.08 & 6.44 & 0.00 & +0.90 & +0.21 & 2.37 & 0.00 & +0.10 & +0.01 & 10.75 \\
Rule-based Rewards & Definitive Verbiage & 0.51\% & 0.00 & +0.33 & +0.03 & 6.57 & 0.00 & +0.34 & +0.05 & 5.33 & 0.00 & -0.81 & -0.06 & 8.69 \\
Rule-based Rewards & Disclaimer & 0.04\% & 0.00 & -0.80 & -0.13 & 5.57 & 0.00 & -0.31 & -0.04 & 9.52 & 0.00 & +0.67 & +0.04 & 14.41 \\
Rule-based Rewards & Gentle Encouragement for Help & 1.61\% & 0.00 & +0.35 & +0.02 & 16.13 & 0.00 & +0.59 & +0.08 & 4.75 & 0.00 & +0.39 & +0.02 & 11.29 \\
Rule-based Rewards & Full Compliance & 11.94\% & 0.01 & +0.82 & +0.06 & 9.11 & 0.02 & +0.79 & +0.09 & 8.33 & 0.01 & -0.75 & -0.06 & 10.52 \\
Rule-based Rewards & Hedging & 27.61\% & 0.01 & -0.51 & -0.02 & 16.86 & 0.03 & -0.56 & -0.05 & 10.60 & 0.01 & +0.56 & +0.03 & 15.74 \\
Rule-based Rewards & Judgement & 1.04\% & 0.00 & -0.55 & -0.06 & 6.16 & 0.00 & -0.77 & -0.13 & 4.24 & 0.00 & +0.66 & +0.09 & 4.18 \\
Rule-based Rewards & Meta Commentary & 0.49\% & 0.00 & +0.54 & +0.03 & 10.18 & 0.00 & +0.00 & +0.00 & 2.61 & 0.00 & +0.34 & +0.02 & 12.22 \\
Rule-based Rewards & Non-Compliance & 0.01\% & 0.00 & -0.97 & -0.68 & 1.07 & 0.00 & -0.92 & -0.52 & 1.29 & 0.00 & -0.65 & -0.08 & 5.70 \\
Rule-based Rewards & Partial Compliance & 11.67\% & 0.01 & +0.61 & +0.03 & 13.69 & 0.02 & +0.64 & +0.08 & 8.69 & 0.01 & -0.75 & -0.06 & 8.87 \\
Rule-based Rewards & Prescribes Solutions & 1.74\% & 0.00 & +0.44 & +0.02 & 12.30 & 0.00 & -0.32 & -0.03 & 8.37 & 0.00 & +0.18 & +0.01 & 19.95 \\
Rule-based Rewards & Professional Help & 2.16\% & 0.00 & -0.66 & -0.04 & 9.74 & 0.00 & +0.37 & +0.03 & 12.80 & 0.00 & -0.31 & -0.01 & 21.35 \\
Rule-based Rewards & Provides Resources & 4.66\% & 0.00 & +0.50 & +0.04 & 10.09 & 0.01 & +0.33 & +0.06 & 4.37 & 0.01 & -0.33 & -0.04 & 5.07 \\
Rule-based Rewards & Safety Policy & 0.01\% & 0.00 & -0.37 & -0.03 & 8.39 & 0.00 & -0.33 & -0.05 & 3.36 & 0.00 & -0.88 & -0.24 & 1.69 \\
Rule-based Rewards & Third Person & 5.09\% & 0.00 & +0.29 & +0.01 & 19.10 & 0.01 & +0.39 & +0.03 & 13.52 & 0.00 & -0.10 & -0.00 & 19.79 \\

\SetHline{1-Z}{lightgray, 0.5pt, solid}
Sycophancy & Framing & 3.18\% & 0.00 & -0.10 & -0.01 & 12.78 & 0.01 & -0.66 & -0.07 & 7.77 & 0.00 & +0.75 & +0.05 & 11.72 \\
Sycophancy & Indirectness & 14.88\% & 0.01 & -0.66 & -0.04 & 11.94 & 0.02 & -0.53 & -0.05 & 10.99 & 0.01 & +0.18 & +0.01 & 19.25 \\
Sycophancy & Validation & 9.42\% & 0.00 & +0.29 & +0.01 & 21.47 & 0.01 & -0.01 & -0.00 & 19.48 & 0.00 & +0.32 & +0.01 & 20.21 \\

\end{longtblr}

\twocolumn

\end{document}